\ifdefined\pdftexversion \pdfoutput=1 \fi

\documentclass[sigconf]{acmart}

\usepackage{booktabs}
\usepackage{graphicx}
\usepackage{algorithm}
\usepackage{algorithmic}
\usepackage{amsmath}
\usepackage{xcolor}
\usepackage{listings}
\usepackage{subcaption}
\usepackage{enumitem}
\usepackage{tcolorbox}
\tcbuselibrary{breakable,skins}
\usepackage[framemethod=tikz]{mdframed}

\newtcolorbox{sopbox}[1][]{
  colback=gray!8,
  colframe=gray!60,
  fonttitle=\bfseries\small,
  title=#1,
  boxrule=0.5pt,
  left=1mm,
  right=1mm,
  top=1mm,
  bottom=1mm,
  fontupper=\small,
  width=\linewidth
}

\newtcolorbox{kbbox}[1][]{
  colback=gray!8,
  colframe=gray!60,
  fonttitle=\bfseries\small,
  title=#1,
  boxrule=0.5pt,
  left=1mm,
  right=1mm,
  top=1mm,
  bottom=1mm,
  fontupper=\small,
  width=\linewidth
}

\newtcolorbox{errorbox}[1][]{
  colback=red!8,
  colframe=red!60,
  fonttitle=\bfseries\small,
  title=#1,
  boxrule=0.5pt,
  left=1mm,
  right=1mm,
  top=1mm,
  bottom=1mm,
  fontupper=\small,
  width=\linewidth
}

\newtcolorbox{analysisbox}[1][]{
  colback=blue!8,
  colframe=blue!50,
  fonttitle=\bfseries\small,
  title=#1,
  boxrule=0.5pt,
  left=1mm,
  right=1mm,
  top=1mm,
  bottom=1mm,
  fontupper=\small,
  width=\linewidth
}

\newtcolorbox{fixbox}[1][]{
  colback=green!8,
  colframe=green!50,
  fonttitle=\bfseries\small,
  title=#1,
  boxrule=0.5pt,
  left=1mm,
  right=1mm,
  top=1mm,
  bottom=1mm,
  fontupper=\small\raggedright,
  boxsep=1mm,
  before upper={\tolerance=9999\emergencystretch=3em}
}

\newtcolorbox{phasebox}[2][]{
  colback=#2,
  colframe=#2,
  fontupper=\bfseries\small\color{white},
  boxrule=0pt,
  arc=2pt,
  left=3mm,
  right=3mm,
  top=1mm,
  bottom=1mm,
  #1
}

\newtcolorbox{policybox}[1][]{
  colback=gray!8,
  colframe=gray!60,
  fonttitle=\bfseries\small,
  title=#1,
  boxrule=0.5pt,
  left=1mm,
  right=1mm,
  top=1mm,
  bottom=1mm,
  fontupper=\small,
  width=\linewidth
}

\newcommand{\systemname}{\textsc{Trace}}
\newcommand{\fullname}{TRajectory Attribution for Context Engineering}
\newcommand{\detector}{\textsc{Detector}}
\newcommand{\reflector}{\textsc{Root Cause}}
\newcommand{\recommender}{\textsc{Recommender}}

\makeatletter
\renewcommand{\@seccntformat}[1]{\csname the#1\endcsname\quad}
\makeatother

\acmConference[KDD '26]{Proceedings of the 32nd ACM SIGKDD Conference on Knowledge Discovery and Data Mining}{August 9--13, 2026}{Jeju, South Korea}
\acmYear{2026}
\copyrightyear{2026}

\renewcommand{\fullname}{TRajectory Attribution for Automated Context Engineering}

\begin{document}

\setcounter{secnumdepth}{3}
\setcounter{tocdepth}{3}

\title{\systemname{}: TRajectory Attribution for Automated Context Engineering}

\author{Yikai Zhao}
\affiliation{%
  \institution{Amazon}
  \city{Dallas}
  \state{TX}
  \country{USA}
}
\email{yikai@amazon.com}

\author{Pradeep Kumar Misra}
\affiliation{%
  \institution{Amazon}
  \city{Dallas}
  \state{TX}
  \country{USA}
}
\email{pkmisra@amazon.com}

\author{Saurabh Pandey}
\affiliation{%
  \institution{Amazon}
  \city{Seattle}
  \state{WA}
  \country{USA}
}
\email{saurabpd@amazon.com}

\begin{abstract}
\emergencystretch=1em
Production AI agents fail when their context sources---system prompts, knowledge bases, tool descriptions, and procedural skills---contain errors or gaps. Current maintenance approaches rely on manual log review and ad-hoc debugging, creating a scalability bottleneck as interaction volume grows.

We present \systemname{} (\fullname{}), an automated feedback loop that mines historical agent trajectories to diagnose and remediate context failures. Our key insight is that agent trajectories are rich with implicit dissatisfaction signals---user corrections, rephrasing patterns, abandonment cues---that reveal precisely where context sources failed, without requiring explicit feedback collection. Unlike model fine-tuning approaches, \systemname{} operates on the context layer, enabling rapid iteration without retraining.

The system makes four contributions: (1) a trajectory mining framework that systematically extracts diagnostic information from historical agent executions; (2) multi-component causal attribution that extends textual gradients from monolithic prompt optimization to heterogeneous context sources (skills, knowledge bases, tools, prompts); (3) exploratory verification where agents actively read context sources to distinguish content gaps requiring CREATE operations from stale content requiring UPDATE---achieving 96\% operation accuracy; and (4) a reusable simulation methodology and verifiable evaluation benchmark addressing the absence of open datasets for context debugging, with a six-category fault taxonomy, complete ground truth annotations, and a cross-layer verification protocol that can be adopted to generate domain-specific benchmarks.

Evaluation on 60 dissatisfaction traces spanning three complexity tiers (up to 16 execution nodes) achieves 72.7\% root cause node attribution and 82\% end-to-end fix effectiveness---demonstrating that over 80\% of context-layer failures could be automatically diagnosed and correctly remediated by mining historical agent trajectories, an overlooked resource in production systems.
\end{abstract}

\ccsdesc[500]{Computing methodologies~Causal reasoning and diagnostics}
\ccsdesc[500]{Computing methodologies~Multi-agent systems}
\ccsdesc[300]{Computing methodologies~Anomaly detection}

\keywords{context engineering, causal attribution, anomaly detection, automated feedback loops, root cause analysis, textual gradients, multi-agent systems}

\maketitle


\section{Introduction}

Modern AI agents are complex systems whose behavior is shaped by carefully engineered \textit{context}---system prompts that define behavior, tool descriptions that guide function calling, knowledge bases that provide domain information through retrieval-augmented generation (RAG), and Skills files that encode multi-step workflows. This practice of \textit{context engineering}~\cite{llmagentsurvey2025} has emerged as the primary mechanism for customizing and improving agent capabilities without retraining the underlying model. However, when users interact with these agents and experience dissatisfaction, the root cause may lie in any of these context sources, making diagnosis challenging and remediation labor-intensive.

Current approaches to maintaining AI agents rely heavily on manual review of interaction logs, ad-hoc error reports, and periodic audits. This creates a fundamental scalability bottleneck: as interaction volume grows, the gap between failures occurring and fixes being applied widens, leading to accumulated user dissatisfaction and degraded agent performance.

To address this challenge, we present \systemname{} (\fullname{}), an automated feedback loop that mines historical agent trajectories to diagnose and remediate context failures. Our approach is grounded in three key insights:

First, historical agent trajectories represent an untapped resource for context engineering. Organizations deploying AI agents generate vast amounts of execution data---not just conversation logs, but complete trajectories including reasoning traces, tool invocations, retrieved contexts, and user reactions. Yet this resource remains largely underutilized for systematic context improvement. While existing approaches use production data for model fine-tuning (RLHF) or require explicit human labels, the implicit signals already present in these trajectories---user corrections, rephrasing patterns, escalation language, semantic drift---provide rich diagnostic information about context failures without additional annotation costs. The DRIFT framework~\cite{drift2025} demonstrates that such dissatisfaction signals occur approximately twice as frequently as explicit satisfaction signals and contain richer semantic information.

Second, conversation trajectories are differentiable. TextGrad~\cite{textgrad2024} introduces the concept of ``textual gradients''---semantic descriptions of how each component in a computational graph should change to improve outcomes. We adapt this concept to conversation trajectories, treating each \textit{decision point} as a node that can receive attribution for the final outcome. A decision point is any step where the agent makes a choice influenced by its context sources: selecting which tool to invoke (guided by tool descriptions), retrieving knowledge (shaped by the KB content and/or embeddings), or formulating a response (constrained by the system prompt).

Third, specialized agents enable modular optimization. The ACE Framework~\cite{ace2024} separates concerns into Generator, Reflector, and Curator roles. We adopt this architecture, allowing each agent to specialize in its task while maintaining clear interfaces between components.

\textbf{Novel Contributions.} This work advances context engineering through four innovations:
\begin{enumerate}
    \item \textbf{Trajectory Mining for Context Engineering}: We present a novel framework for systematically mining historical agent trajectories---an overlooked resource---to diagnose and remediate context failures. While prior work (e.g., DRIFT~\cite{drift2025}) uses implicit dissatisfaction signals for model fine-tuning, we apply them to the context layer, enabling rapid iteration without model retraining.

    \item \textbf{Implicit Signal Exploitation}: We exploit implicit user feedback signals---corrections, rephrasing patterns, abandonment cues---that typically go unnoticed in production logs. These signals reveal precisely where context sources are failing without requiring explicit feedback collection or annotation.

    \item \textbf{End-to-End Attribution with Exploratory Verification}: We extend textual gradients from monolithic prompt optimization to multi-component context sources (prompts, KB, tools, Skills), and introduce exploratory verification where the \recommender{} agent actively reads context source files before generating recommendations. This verification is essential: to recommend the correct remediation---CREATE (content missing) vs.\ UPDATE (content exists but wrong)---the agent must verify whether the content actually exists. Without exploration, the agent achieves only 33\% accuracy on this decision; with exploration, accuracy rises to 83\%, a 50 percentage point improvement.

    \item \textbf{Verifiable Evaluation Benchmark and Simulation Methodology}: To our knowledge, no open-source dataset or industry-standard benchmark exists for context engineering debugging from agent trajectories. While \systemname{} was developed and validated on a proprietary enterprise agent system, confidentiality constraints preclude disclosure of production data. We therefore introduce a systematic three-layer simulation methodology for generating synthetic agent traces with perfect ground truth. Our benchmark provides: (i) a six-category fault taxonomy covering common context engineering failures as a sample, (ii) traces of varying complexity (2--16 nodes) with documented cascade effects, (iii) complete attribution ground truth, and (iv) a cross-layer verification protocol ensuring dataset integrity. Critically, the simulation methodology itself is a contribution: we provide detailed procedures (Appendix~\ref{app:simulation}) that can be adopted as a reusable skill for GenAI agents to generate domain-specific evaluation datasets, enabling the community to extend this benchmark to new domains and fault types.
\end{enumerate}

\section{Related Work}

\systemname{} builds upon four research streams: textual gradients for weight-free optimization, dissatisfaction-driven learning, agentic self-improvement architectures, and knowledge base maintenance.

\subsection{Textual Gradients and Prompt Optimization}

The concept of optimizing LLM behavior without weight updates has gained traction through frameworks that treat prompts as differentiable variables. \textbf{TextGrad}~\cite{textgrad2024} introduced ``textual gradients''---semantic descriptions of how each component in a computational graph should change to improve outcomes. TextGrad implements a \texttt{backward()} pass that propagates semantic feedback via meta-prompting, enabling Textual Gradient Descent (TGD) optimization. Zhou et al.~\cite{zhou2024semantic} analyze TextGrad's backward pass and note that it processes each predecessor variable independently---appropriate for optimization where variables are updated separately. \textbf{ProTeGi}~\cite{protegi2023} treats prompt optimization as beam search to address LLM output variance. \textbf{CRISPO}~\cite{crispo2025} decomposes critique into aspect-specific signals for surgical updates. \textbf{PACE}~\cite{pace2023} adopts an Actor-Critic architecture to prevent ``Prompt Drift.''

\subsection{Dissatisfaction-Driven Learning}

Traditional RLHF~\cite{rlhf2022} relies on explicit preference signals, which are sparse and expensive to collect. The \textbf{DRIFT} framework~\cite{drift2025} demonstrates that implicit dissatisfaction signals---conversational repair, refinement trajectories, semantic drift---occur twice as frequently as satisfaction signals and contain richer diagnostic information. \textbf{Inverse Constitutional AI (ICAI)}~\cite{icai2025} extracts generalizable principles from pairwise corrections, reverse-engineering user preferences from conversation traces. Meanwhile, Reinforcement Learning with Verifiable Rewards (RLVR) methods leverage verifiable outcomes (e.g., code execution results, tool errors) as reward signals without human labeling. Both DRIFT and RLVR target model weight optimization through fine-tuning.

\subsection{Agentic Self-Improvement Architectures}

The \textbf{Agentic Context Engineering (ACE)} framework~\cite{ace2024} establishes a tripartite design pattern---Generator, Reflector, Curator---that separates task execution from system optimization. The Generator executes tasks using a dynamic ``Playbook''; the Reflector performs root cause analysis on failures; the Curator manages Playbook updates through Delta Operations (ADD, UPDATE, DELETE) to prevent context collapse. Research demonstrates that maintaining context as a dynamic artifact improves agentic benchmark performance by over 10\%~\cite{ace2024}. The literature on LLM-based AI agents~\cite{llmagentsurvey2025} surveys feedback mechanisms including environmental feedback (tool results), human feedback (corrections), and model feedback (self-critique).

\subsection{Knowledge Base and Memory Management}

For RAG systems, optimization often requires repairing the external knowledge base. \textbf{Mem0}~\cite{mem02025} introduces structured memory management with explicit CRUD capabilities and conflict detection---performing semantic comparison before deciding to ADD, UPDATE, MERGE, or DELETE entries. \textbf{Amber}~\cite{amber2025} addresses retrieval failures by optimizing the retrieval process through an Adaptive Information Collector that learns mappings between vague queries and successful refined queries.

A detailed positioning of \systemname{} relative to each of these four streams---attribution vs.\ optimization, context layer vs.\ model weights, active exploration vs.\ passive acceptance, and the diagnostic layer for KB---is provided in Appendix~\ref{app:positioning}.

\section{System Architecture}

\begin{figure*}[t]
    \centering
    \includegraphics[width=0.85\textwidth]{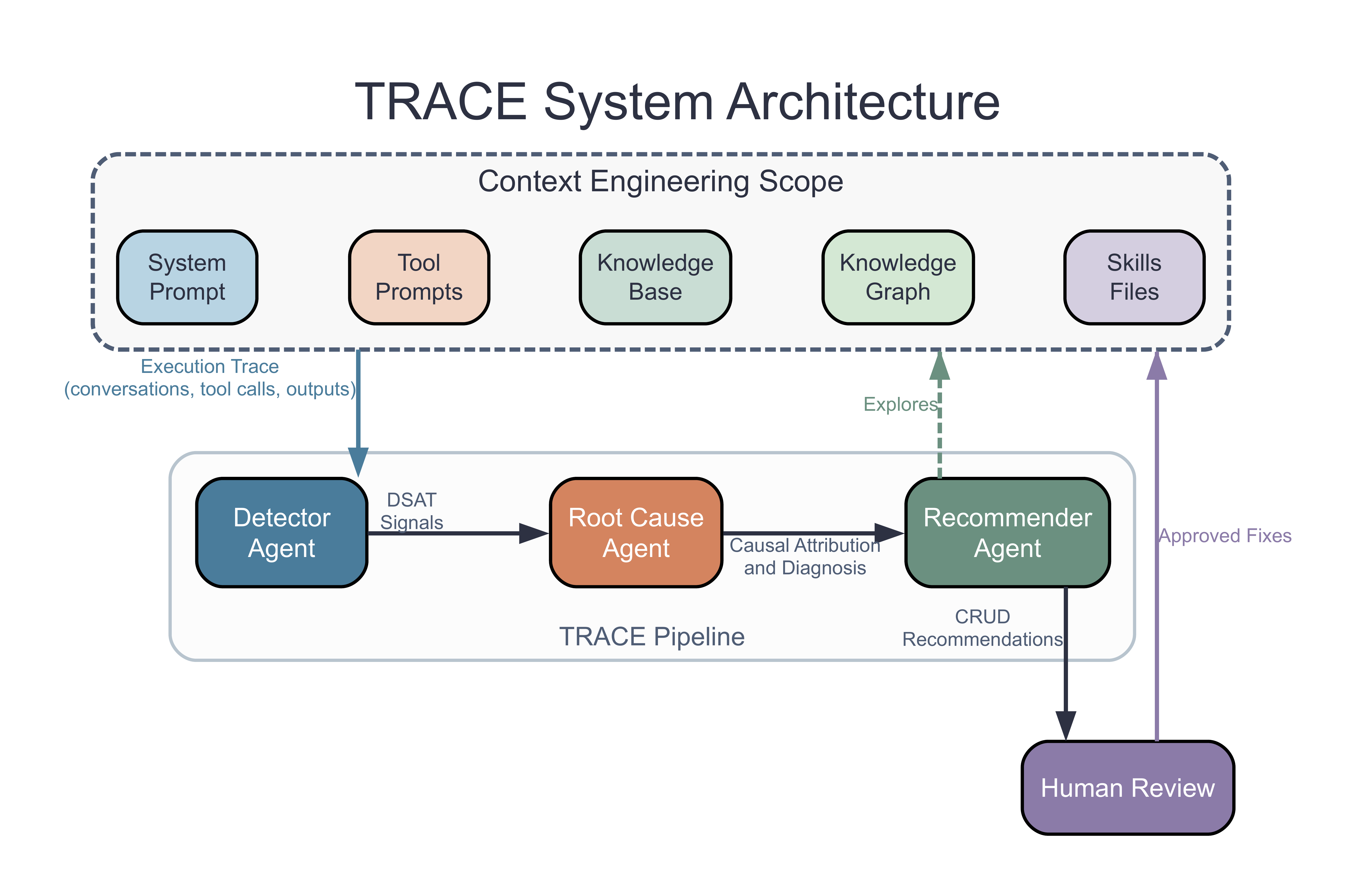}
    \caption{Overview of the \systemname{} architecture. The system consists of three specialized agents: \detector{} identifies dissatisfaction signals from agent trajectories, \reflector{} performs holistic attribution to identify root causes, and \recommender{} generates CRUD recommendations for human review. Approved fixes are applied to the chatbot system's context engineering scope---system prompt, tool prompts, knowledge base, knowledge graph, or Skill files.}
    \label{fig:architecture}
\end{figure*}

Figure~\ref{fig:architecture} presents an overview of the \systemname{} architecture. The system operates as a feedback loop that continuously monitors agent trajectories, detects failures, attributes root causes, and generates recommendations for human review. A per-component overview is in Appendix~\ref{app:component-overview}.

We now detail each agent's design and implementation. A complete end-to-end worked example demonstrating the full pipeline on a realistic failure scenario is provided in Appendix~\ref{app:example}.

\section{Detector Agent}
\label{sec:detector}

The \detector{} Agent identifies dissatisfaction signals from agent trajectories, distinguishing between explicit signals (direct feedback) and implicit signals (linguistic cues). The \detector{} requires structured agent trajectories---complete execution records, not just user/assistant messages (full input requirements in Appendix~\ref{app:detector-inputs}).

\subsection{Signal Detection}

The \detector{} takes as input a complete conversation segment---including the user's original query, the agent's response, and any follow-up exchanges where the user provides corrections or expresses dissatisfaction. Through qualitative inspection of agent conversation traces, we identified recurring patterns of user dissatisfaction that fall into two categories: \textit{explicit signals} (e.g., thumbs-down feedback, tool execution errors, session abandonment) and \textit{implicit signals} (e.g., conversational repair, repetition, negative sentiment, semantic drift). These signal types are embedded in the detection prompt, enabling the LLM to scan conversations and flag sessions warranting further analysis.

The \detector{} outputs a binary DSAT determination along with a confidence score. Sessions exceeding the confidence threshold (or containing explicit signals like tool errors) are escalated to the \reflector{} for root cause attribution. The complete signal taxonomy is detailed in Appendix~\ref{app:taxonomy}, and the detection prompt is provided in Appendix~\ref{app:prompts}.


\section{Root Cause Agent}
\label{sec:reflector}

The \reflector{} Agent analyzes the conversation trajectory to identify where and why the agent failed, generating textual gradients that attribute responsibility to specific components.

\subsection{Trajectory as Context Graph}

We model the conversation as a graph where each node represents a context source that influenced the agent's response (full formal definition, equation, and figure in Appendix~\ref{app:trajectory-graph}). Each node $v$ has associated content $c_v$ and output $o_v$. This graph structure provides a conceptual framework for understanding how context flows through the agent, though our primary attribution method processes all nodes holistically rather than traversing the graph. The \reflector{} attributes failures to one of six root cause categories spanning the context engineering surface (detailed in Section~\ref{sec:fault-taxonomy}).

\subsection{Delta-Guided Holistic Attribution}

Our primary attribution method presents the complete trajectory to the LLM in a single pass. The \textit{delta}---the discrepancy between what the user expected and what the agent produced---serves as the ``loss signal'' that guides attribution. The \reflector{} reads only the agent's execution trace---chain-of-thought, tool calls, and tool outputs---and does not separately inspect raw context files; the chain-of-thought is the key diagnostic signal, since the agent's own reasoning names the sources that shaped each decision (e.g., \textit{``According to the KB \ldots''} or \textit{``Following the SOP for \ldots''}), letting the \reflector{} identify candidate root-cause sources by reading the trace alone. Verification of those candidate files is deferred to the \recommender{} stage. The LLM is instructed to reason over the trace in \textit{reverse temporal order}---gradient-descent style: starting from the final response (where the loss is observed), it walks backward (response $\rightarrow$ tool outputs $\rightarrow$ thinking $\rightarrow$ inputs) to locate the earliest node whose content is inconsistent with the delta. This mirrors how textual gradients~\cite{textgrad2024} propagate semantic feedback backward through a computational graph, condensed into a single LLM call.

\begin{algorithm}[t]
\caption{Holistic Attribution (Primary Method)}
\label{alg:holistic}
\begin{algorithmic}[1]
\REQUIRE Trace $\mathcal{T} = (v_1, v_2, \ldots, v_n)$: time-ordered sequence of nodes from the agent's execution---chain-of-thought, tool calls, tool outputs---each with input $c_v$ and output $o_v$.
\REQUIRE User correction $u_{\text{corr}}$, Agent response $r_{\text{agent}}$
\ENSURE Root cause node $v^*$, fault category $c^*$

\STATE \textbf{// Step 1: Compute delta (loss signal)}
\STATE $\delta \gets \text{ExtractDelta}(u_{\text{corr}}, r_{\text{agent}})$
\STATE \textit{// $\delta$ captures ``Expected X but got Y''}

\STATE \textbf{// Step 2: Build reverse-ordered evidence bundle}
\STATE $\mathcal{T}_{\text{rev}} \gets (v_n, v_{n-1}, \ldots, v_1)$ \textit{// response$\to\cdots\to$query}
\STATE $\mathcal{C} \gets \big[(v, c_v, o_v)\big]_{v \in \mathcal{T}_{\text{rev}}}$ \textit{// ordered list following $\mathcal{T}_{\text{rev}}$}
\STATE \textit{// Trace only; raw context files are NOT separately inspected here}

\STATE \textbf{// Step 3: Single-pass backward attribution}
\STATE $(v^*, c^*) \gets \text{LLM}(\textsc{HolisticPrompt}, \delta, \mathcal{C})$
\STATE \textit{// LLM walks $\mathcal{T}_{\text{rev}}$ from response backward, using the agent's}
\STATE \textit{// chain-of-thought references to localize the source of $\delta$}

\STATE \textbf{return} $(v^*, c^*)$
\end{algorithmic}
\end{algorithm}

The simultaneous view (Algorithm~\ref{alg:holistic}) lets the LLM disambiguate ties between candidate root-cause nodes; the reverse ordering biases reasoning toward the earliest node that introduced the delta rather than later nodes that merely propagated it. This single-call design reduces cost by 16$\times$ versus iterative per-node approaches and avoids information loss from context truncation; we validate it against an iterative baseline in Appendix~\ref{app:ablations}. The reasoning traces themselves are diagnostically rich: the \reflector{} extracts references like ``According to the KB...'' $\rightarrow$ check that KB entry, or ``Following the SOP for...'' $\rightarrow$ check that Skill file, enabling precise attribution even when the error's surface manifestation is distant from its root cause.

\section{Recommender Agent}
\label{sec:recommender}

The \recommender{} Agent represents a critical departure from conventional pipeline architectures: rather than simply consuming the \reflector{}'s output and generating fixes, it conducts its own comprehensive exploration of the agent's context sources before producing recommendations. This section details this novel \textit{deep exploration methodology} and the justification analysis framework that distinguishes actionable issues from noise.

\subsection{Independent Exploration: Hypothesis Verification, Not Passive Acceptance}

A fundamental design principle of \systemname{} is that the \recommender{} treats the \reflector{}'s root cause analysis as a \textit{hypothesis to verify}, not a conclusion to accept. This distinction is crucial for several reasons:

\begin{enumerate}
    \item \textbf{Attribution may be incomplete.} The \reflector{}'s holistic attribution identifies the component with highest attribution, but related issues in other components may also require attention. An outdated KB entry might indicate that multiple related entries need updating.

    \item \textbf{Context sources interact.} A failure attributed to a KB entry might actually stem from a system prompt that should have instructed the agent to verify KB freshness. The \recommender{} explores these interactions.

    \item \textbf{Ground truth requires verification.} The \reflector{} identifies \textit{what} component contributed to failure; the \recommender{} must verify \textit{whether} that component is actually incorrect by cross-referencing authoritative sources.
\end{enumerate}

\textbf{Exploration Tools.} The \recommender{} has access to the same context sources as the production agent, enabling comprehensive investigation. The full set of representative tools (e.g., \texttt{search\_kb}, \texttt{read\_kb\_entry}, \texttt{list\_skills}, \texttt{read\_skill}, \texttt{read\_system\_prompt}, \texttt{read\_tool\_prompt}) is given in Appendix~\ref{app:exploration}.

\textbf{Exploration Protocol.} Upon receiving a \texttt{RootCauseAnalysis}, the \recommender{} executes a multi-phase exploration---\textit{read} the implicated component, \textit{search} authoritative sources, \textit{cross-reference} to validate, and \textit{explore} related components (full procedure in Appendix~\ref{app:exploration}).

This exploration may \textit{confirm} the \reflector{}'s attribution, \textit{refine} it with additional context, \textit{expand} it to include related issues, or in some cases \textit{override} it when exploration reveals the true root cause lies elsewhere.

Based on exploration findings, the \recommender{} outputs structured CRUD recommendations specifying the operation (CREATE, UPDATE, DELETE, or NO\_ACTION), target path, recommended change, and supporting evidence. When authority signals conflict, the issue is escalated for human resolution. The complete diagnostic framework, output schema, and prompt template are provided in Appendix~\ref{app:prompts}.

\section{Evaluation Setup}
\label{sec:experiments}

We evaluate \systemname{} on a synthetic dataset of agent conversation traces designed to simulate realistic failure modes encountered in enterprise AI agent deployments.

\textbf{Why Synthetic Data?} Three factors motivated our simulation-based evaluation (full discussion in Appendix~\ref{app:eval-rationale}).

Our simulation methodology addresses these challenges by generating traces with verifiable ground truth while maintaining structural realism. Importantly, we contribute this methodology as a reusable framework (Appendix~\ref{app:simulation}) that practitioners can adapt to generate domain-specific benchmarks.

\subsection{Dataset Description}

Our evaluation dataset consists of 75 conversation traces:
\begin{itemize}[nosep,leftmargin=*]
    \item \textbf{60 DSAT traces}: Conversations containing user dissatisfaction signals with known root causes
    \item \textbf{15 control traces}: Successful conversations without failures (negative examples)
\end{itemize}

Each trace consists of a sequence of \textit{nodes}, where each node represents a discrete step in the agent's execution: a skill file lookup, KB search, database query, or response generation. Trace complexity is measured by node count---more nodes mean longer reasoning chains where the root cause may be obscured by subsequent processing steps and cascade effects. The DSAT traces span three complexity tiers (simple: 2--8 nodes, complex: 9--11 nodes, difficult: 15--16 nodes) and six fault categories. The complete dataset composition is detailed in Appendix~\ref{app:simulation}.

\subsection{Evaluation Benchmark: Verifiable Context Debugging Dataset}

Our benchmark implements a three-layer generative architecture comprising Context Sources, Fault Definitions, and Execution Traces (full layer descriptions in Appendix~\ref{app:eval-rationale}; the simulation framework is detailed in Appendix~\ref{app:simulation}).

\subsubsection{Fault Taxonomy}
\label{sec:fault-taxonomy}

Our taxonomy covers six failure modes spanning the context engineering surface: \texttt{SKILL\_FILE\_STALE} (n=12), \texttt{KB\_CONTENT\_STALE} (n=12), \texttt{KB\_CONTENT\_GAP} (n=11), \texttt{TOOL\_PROMPT\_ERROR} (n=9), \texttt{SYSTEM\_PROMPT\_GAP} (n=6), and \texttt{RETRIEVAL\_FAILURE} (n=10); per-category descriptions are in Appendix~\ref{app:fault-taxonomy}. This taxonomy distinguishes \textit{presence faults} (incorrect information exists) from \textit{absence faults} (correct information missing)---a distinction the \recommender{}'s exploration stage resolves through active verification.

\subsubsection{Ground Truth and Verification}

Each trace includes structured ground truth: root cause node ID with component path and expected CRUD recommendation. We implement a five-point verification protocol ensuring tool outputs match source files, fault annotations exist at referenced locations, and cross-references are consistent. All 75 traces pass verification. Appendix~\ref{app:simulation} provides the full ground truth schema, verification procedures, and reproducibility guide.

\subsection{Evaluation Protocol}

We evaluate each pipeline component independently and measure end-to-end accuracy. All metrics are computed on the 60 DSAT traces with ground truth labels.

\textbf{\detector{} Evaluation}: Binary classification measuring whether the agent correctly identifies DSAT vs.\ non-DSAT traces. We report precision, recall, and F1 score. The 15 control traces serve as negative examples.

\textbf{\reflector{} Agent Evaluation}: Root cause attribution measured by \textit{Node Accuracy} (the fraction of traces where the predicted root cause node ID matches the ground truth---our primary metric) and \textit{Component Type Accuracy} (the fraction of traces where the predicted component type matches ground truth). Per-metric definitions are in Appendix~\ref{app:eval-rationale}. We intentionally de-emphasize fine-grained fault category classification (e.g., KB\_STALE vs KB\_GAP) at this stage. From the conversation alone, distinguishing ``outdated information'' from ``missing information'' is often ambiguous---both manifest as the user providing a correction. This distinction is better resolved at the \recommender{} stage through active exploration of context sources.

\textbf{\recommender{} Evaluation}: We measure two aspects of recommendation quality: \textit{Operation Accuracy} (predicted CRUD operation matches ground truth) and \textit{Path Accuracy} (predicted target file path matches ground truth). Per-metric definitions are in Appendix~\ref{app:eval-rationale}.

\textbf{End-to-End Evaluation}: We report component-level accuracies and compute fix effectiveness as the fraction of traces where the full pipeline produces a correct, actionable recommendation (correct operation AND correct target path). We also analyze error cascades---how errors in upstream components (e.g., wrong node attribution) affect downstream recommendations. Formal metric definitions are provided in Appendix~\ref{app:metrics}.

\section{Results}
\label{sec:results}

\subsection{\detector{} Performance}

Both \detector{} and a vanilla LLM baseline achieve perfect binary DSAT detection (precision=1.0, recall=1.0, F1=1.0) on our evaluation dataset. The \detector{} uses a structured prompt embedding our eight-category DSAT signal taxonomy with concrete examples (e.g., distinguishing CORRECTION from CONVERSATIONAL\_REPAIR), while the vanilla baseline uses a simple prompt asking ``Is the user dissatisfied? Output true/false.'' without any taxonomy guidance.

The equivalent performance reveals an important insight: modern foundation models possess strong inherent ability to recognize user dissatisfaction from conversational cues, regardless of whether they receive explicit signal taxonomies. The taxonomy-guided \detector{} provides richer diagnostic information (specific signal types, evidence quotes, trigger turns) that benefits downstream attribution, but for binary DSAT/non-DSAT classification, the additional structure provides no accuracy improvement.

\subsection{\reflector{} Agent Performance}

\begin{table}[h]
\centering
\small
\begin{tabular}{lcc}
\toprule
\textbf{Metric} & \textbf{Value} & \textbf{95\% CI} \\
\midrule
Node Accuracy (Acc@1) & 72.7\% & [59\%, 85\%] \\
Acc@3 & 85.0\% & [73\%, 95\%] \\
\bottomrule
\end{tabular}
\caption{\reflector{} root cause attribution accuracy with 95\% bootstrap CIs (n=60). Node accuracy measures exact match; Acc@3 measures whether the correct node is among top-3 attributed.}
\label{tab:reflector}
\end{table}

\textbf{Key findings}: (1) Overall node accuracy of 72.7\% validates the delta-guided holistic attribution approach---the \reflector{} correctly identifies where in the trajectory the error originated in nearly three-quarters of cases. (2) Acc@3 of 85\% shows strong ranking quality: even when the exact node is missed, the correct answer is typically among the top-3 attributed nodes. A complementary per-component-type accuracy breakdown---which groups fault categories by component rather than distinguishing STALE vs.\ GAP---and three additional findings on Skill/Prompt, KB, and Tool components are reported in Appendix~\ref{app:component-type}.

\textbf{Note}: \reflector{} is evaluated on NODE accuracy and COMPONENT TYPE only. The fine-grained STALE vs GAP distinction is evaluated at the \recommender{} stage, where tools enable verification of content existence.

\subsection{\recommender{} Performance}

\begin{table}[h]
\centering
\small
\begin{tabular}{lcc}
\toprule
\textbf{Metric} & \textbf{Value} & \textbf{95\% CI} \\
\midrule
Operation Accuracy & 96\% & [87\%, 100\%] \\
Path Accuracy & 82\% & [69\%, 92\%] \\
\bottomrule
\end{tabular}
\caption{\recommender{} performance with 95\% bootstrap CIs (n=60). Operation accuracy measures correct CRUD operation (CREATE vs UPDATE); Path accuracy measures correct target file identification.}
\label{tab:recommender}
\end{table}

The \recommender{} achieves 96\% operation accuracy, correctly selecting CREATE for content gaps and UPDATE for stale content in nearly all cases. Path accuracy of 82\% reflects the challenge of pinpointing the exact file when multiple KB documents are relevant. The gap between operation (96\%) and path (82\%) accuracy indicates that identifying \textit{what action} to take is easier than identifying \textit{where} to apply it.

\subsection{End-to-End Pipeline}

\begin{table}[h]
\centering
\small
\begin{tabular}{lcc}
\toprule
\textbf{Metric} & \textbf{Value} & \textbf{95\% CI} \\
\midrule
Node Attribution & 72.7\% & [59\%, 85\%] \\
Fix Effectiveness & 82\% & [69\%, 92\%] \\
\bottomrule
\end{tabular}
\caption{End-to-end pipeline accuracy with 95\% bootstrap CIs (n=60). Node Attribution measures correct root cause identification; Fix Effectiveness measures correct CRUD operation AND target path.}
\label{tab:e2e}
\end{table}

\textbf{Metric Interpretation.} Node Attribution (72.7\%) measures whether \systemname{} correctly identifies \textit{where} in the trajectory the failure originated. Fix Effectiveness (82\%) is the stricter end-to-end metric requiring both the correct remediation action (CREATE vs UPDATE) and the correct target file path. The 82\% fix effectiveness demonstrates that mining historical trajectories with \systemname{} could automatically diagnose and correctly remediate over four-fifths of context-layer failures, with the \recommender{}'s exploration recovering from upstream attribution errors 67\% of the time (Table~\ref{tab:ablation2}, Appendix~\ref{app:ablations}). Two ablation studies validating the holistic vs.\ iterative attribution choice and the active exploration vs.\ passive acceptance choice are reported in Appendix~\ref{app:ablations}.


\section{Conclusion}

We presented \systemname{} (\fullname{}), an automated diagnostic system for root cause attribution and context remediation in AI agent deployments. The system addresses a critical scalability challenge in production environments: the labor-intensive process of identifying why agents fail and determining what context sources require modification.

\textbf{Key Results.} Evaluation on 60 DSAT traces across 6 fault categories demonstrates: (1) 72.7\% root cause node accuracy using holistic delta-guided attribution, (2) 82\% end-to-end fix effectiveness with 96\% CRUD operation correctness, and (3) exploration enables 83\% vs 33\% operation accuracy on KB content faults where GAP/STALE distinction is critical.

\textbf{Summary of Technical Innovations.} Our contributions advance the state of the art in context engineering (full point-by-point write-ups in Appendix~\ref{app:innovations}).

\textbf{Impact and Future Directions.} The \systemname{} architecture enables organizations to transition from reactive, manual debugging to proactive, automated improvement cycles. The techniques introduced---particularly delta-guided holistic attribution and exploratory verification---are applicable to any LLM-based agent system where failures stem from multiple interacting context sources. By mining historical trajectories, \systemname{} provides a foundation for self-improving AI systems that learn from operational failures. Future work could extend single-session analysis with cross-session pattern detection to aggregate recurring failure modes across user populations.


\bibliographystyle{ACM-Reference-Format}



\appendix
\setcounter{secnumdepth}{3}  

\section{Detailed Positioning}
\label{app:positioning}

\systemname{} differs from prior work along four dimensions:

\textbf{Attribution vs.\ Optimization.} TextGrad and related frameworks perform \textit{optimization}---iteratively improving variables through textual feedback, processing each variable independently. \systemname{} addresses a different problem: \textit{causal attribution}---identifying which component caused a failure. Attribution requires comparing multiple context sources (skills, KB, tools, prompts) simultaneously to determine which introduced the error. Our holistic approach presents the agent-accessed sources from the trajectory to the LLM in a single pass, achieving equivalent accuracy to iterative methods with 16$\times$ fewer LLM calls.

\textbf{Context Layer vs.\ Model Weights.} DRIFT and RLVR use dissatisfaction signals and verifiable outcomes to optimize model weights through fine-tuning. \systemname{} targets the \textit{context layer}---prompts, knowledge bases, tool definitions, and skills---enabling rapid iteration without retraining. We unify both explicit signals (tool failures, thumbs-down feedback) and implicit signals (corrections, rephrasing patterns) for context-layer attribution.

\textbf{Active Exploration vs.\ Passive Acceptance.} ACE's Curator generates recommendations based on the Reflector's analysis. \systemname{}'s \recommender{} extends this role by actively \textit{exploring} context sources before generating recommendations---reading files, searching the KB, and cross-referencing content to verify hypotheses rather than passively accepting attributed causes. This exploration is critical for distinguishing content gaps (requiring CREATE) from stale content (requiring UPDATE).

\textbf{Diagnostic Layer for KB.} While Mem0 and Amber focus on memory maintenance operations, \systemname{} provides the upstream \textit{diagnostic layer} that determines whether a failure stems from content gaps, content staleness, or retrieval failures---each requiring different remediation strategies. This diagnosis must occur before invoking appropriate CRUD operations.

\section{Detector Input Requirements}
\label{app:detector-inputs}

The \detector{} requires structured agent trajectories---complete execution records, not just user/assistant messages. This includes:

\begin{itemize}
    \item User messages and assistant responses
    \item Thinking/reasoning traces (chain-of-thought)
    \item Tool calls with inputs, outputs, and execution status
    \item KB retrievals with queries and similarity scores
    \item Skill file lookups with paths and sections read
\end{itemize}

\section{Recommender Exploration Tools and Protocol}
\label{app:exploration}

\textbf{Exploration Tools.} The \recommender{} has access to the same context sources as the production agent, enabling comprehensive investigation. The following tools are \textit{representative examples}; the actual tool set may vary depending on the specific context sources available in each deployment:

\begin{itemize}
    \item \texttt{search\_kb(query)}: Search the knowledge base for related content
    \item \texttt{read\_kb\_entry(id)}: Read full KB entry content and metadata
    \item \texttt{list\_skills(path)}: Browse Skills file hierarchy
    \item \texttt{read\_skill(path)}: Read Skill SOP documents
    \item \texttt{read\_system\_prompt()}: Get current system prompt
    \item \texttt{read\_tool\_prompt(name)}: Get tool descriptions
\end{itemize}

\textbf{Exploration Protocol.} Upon receiving a \texttt{RootCauseAnalysis}, the \recommender{} executes a multi-phase exploration:

\begin{enumerate}
    \item \textit{Read the implicated component} to understand its current state and metadata (last modified date, author, version).

    \item \textit{Search for authoritative sources} that should govern the implicated content (e.g., policy documents, official specifications, upstream data sources).

    \item \textit{Cross-reference and validate} by comparing the implicated content against authoritative sources to confirm whether a discrepancy exists.

    \item \textit{Explore related components} that might be affected by the same issue or require coordinated updates.
\end{enumerate}

\section{Evaluation Rationale and Detailed Protocol}
\label{app:eval-rationale}

\subsection{Why Synthetic Data?}

Three factors motivated our simulation-based evaluation:

\begin{enumerate}[nosep,leftmargin=*]
    \item \textbf{Proprietary constraints}: \systemname{} was developed for a production enterprise agent system. Confidentiality requirements preclude disclosure of real interaction data, context sources, or system architecture details.

    \item \textbf{Absence of open benchmarks}: To our knowledge, no open-source dataset exists for evaluating context engineering debugging from agent trajectories. Existing agent evaluation benchmarks (e.g., AgentBench~\cite{liu2023agentbench}, GAIA~\cite{mialon2023gaia}) focus on task completion rather than fault attribution. RAG evaluation datasets (e.g., RGB~\cite{chen2024benchmarking}) evaluate retrieval quality but not end-to-end context debugging with CRUD recommendations.

    \item \textbf{Ground truth requirements}: Rigorous evaluation of root cause attribution requires perfect ground truth---knowing exactly which node introduced each error. Production logs lack such annotations, and manual labeling is prohibitively expensive and subjective.
\end{enumerate}

Our simulation methodology addresses these challenges by generating traces with verifiable ground truth while maintaining structural realism. Importantly, we contribute this methodology as a reusable framework (Appendix~\ref{app:simulation}) that practitioners can adapt to generate domain-specific benchmarks.

\subsection{Evaluation Benchmark: Verifiable Context Debugging Dataset}

Our benchmark implements a three-layer generative architecture (detailed in Appendix~\ref{app:simulation}): (1) \textbf{Context Sources}---23 files across skills, KB, database schemas, and tool definitions with embedded faults; (2) \textbf{Fault Definitions}---structured injections specifying faulty content, correct values, and triggering queries; (3) \textbf{Execution Traces}---75 traces where tool outputs are \textit{exact copies} from context sources, enabling deterministic verification.

\subsection{Detailed Evaluation Protocol}

We evaluate each pipeline component independently and measure end-to-end accuracy. All metrics are computed on the 60 DSAT traces with ground truth labels.

\textbf{\detector{} Evaluation}: Binary classification measuring whether the agent correctly identifies DSAT vs.\ non-DSAT traces. We report precision, recall, and F1 score. The 15 control traces serve as negative examples.

\textbf{\reflector{} Agent Evaluation}: Root cause attribution measured by:
\begin{itemize}[nosep,leftmargin=*]
    \item \textit{Node Accuracy}: The fraction of traces where the predicted root cause node ID matches the ground truth node ID. This is our primary metric---it measures whether the agent correctly identifies \textit{where} in the trajectory the error originated.
    \item \textit{Component Type Accuracy}: The fraction of traces where the predicted component type (skill, KB, tool, or prompt) matches ground truth, regardless of the specific node. This relaxed metric captures cases where the agent identifies the correct \textit{type} of problem even if the exact node differs.
\end{itemize}
We intentionally de-emphasize fine-grained fault category classification (e.g., KB\_STALE vs KB\_GAP) at this stage. From the conversation alone, distinguishing ``outdated information'' from ``missing information'' is often ambiguous---both manifest as the user providing a correction. This distinction is better resolved at the \recommender{} stage through active exploration of context sources.

\textbf{\recommender{} Evaluation}: We measure two aspects of recommendation quality:
\begin{itemize}[nosep,leftmargin=*]
    \item \textit{Operation Accuracy}: The fraction of traces where the predicted CRUD operation (CREATE, UPDATE, DELETE, or NO\_ACTION) matches ground truth. This metric implicitly evaluates GAP vs.\ STALE distinction: GAP faults require CREATE while STALE faults require UPDATE.
    \item \textit{Path Accuracy}: The fraction of traces where the predicted target file path exactly matches ground truth. This metric evaluates whether the recommendation pinpoints the correct file for editing.
\end{itemize}

\textbf{End-to-End Evaluation}: We report component-level accuracies and compute fix effectiveness as the fraction of traces where the full pipeline produces a correct, actionable recommendation (correct operation AND correct target path). We also analyze error cascades---how errors in upstream components (e.g., wrong node attribution) affect downstream recommendations. Formal metric definitions are provided in Appendix~\ref{app:metrics}.

\section{Trajectory as Context Graph}
\label{app:trajectory-graph}

\begin{figure}[h]
    \centering
    \includegraphics[width=0.95\columnwidth]{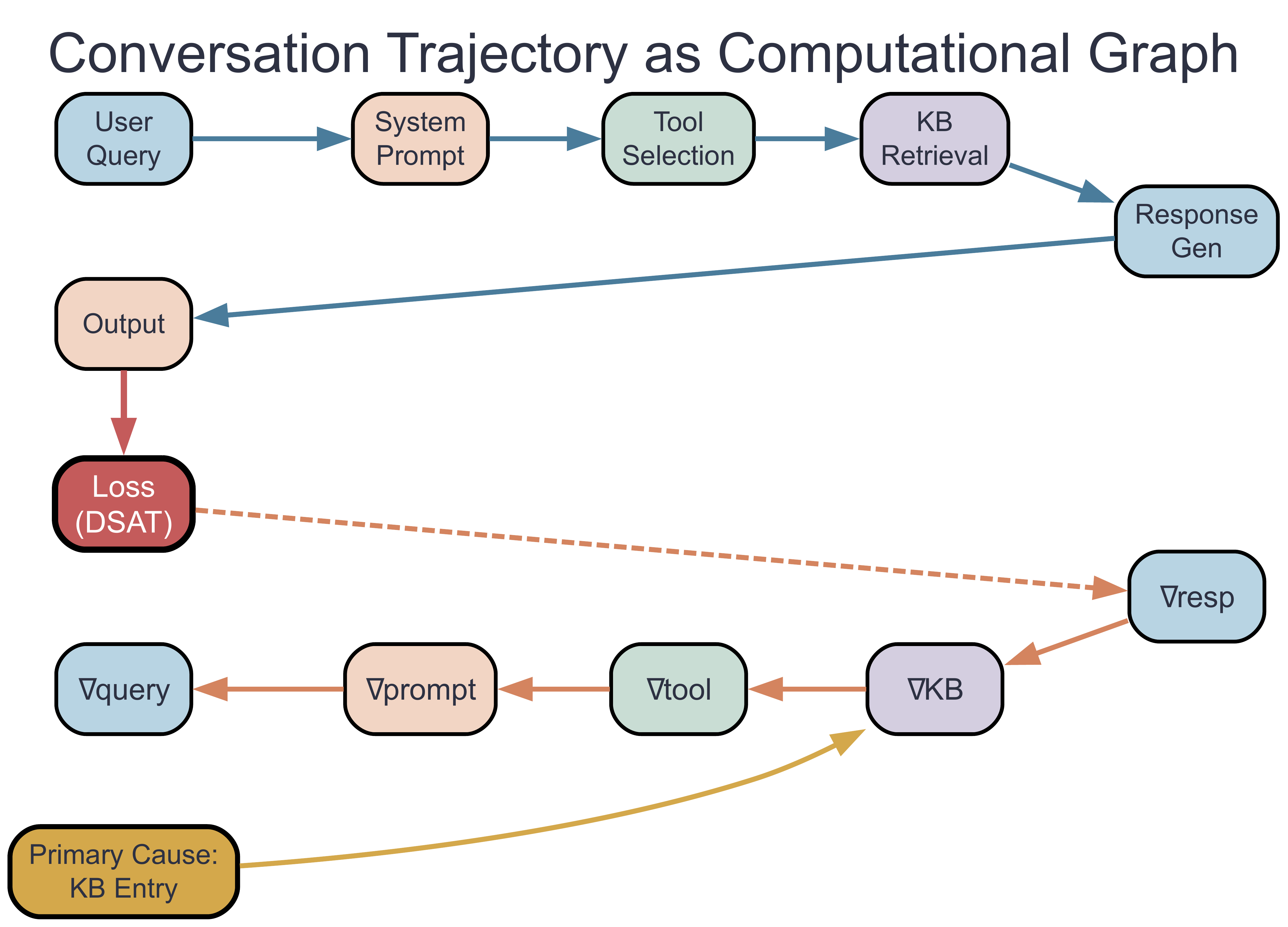}
    \caption{Conversation trajectory modeled as a context graph. Nodes represent context sources (system prompt, KB entries, skill files, tool outputs) that influenced the agent's response. The graph structure captures dependencies for visualization and analysis.}
    \label{fig:trajectory}
\end{figure}

We model the conversation as a graph where each node represents a context source that influenced the agent's response (Figure~\ref{fig:trajectory}):

\begin{equation}
    \mathcal{G} = (V, E), \quad V = \{v_{\text{query}}, v_{\text{prompt}}, v_{\text{tool}}, v_{\text{kb}}, v_{\text{skill}}, v_{\text{resp}}\}
\end{equation}

Each node $v$ has associated content $c_v$ (the prompt text, retrieved chunks, etc.) and output $o_v$ (the decision or generation at that step). This graph structure provides a conceptual framework for understanding how context flows through the agent, though our primary attribution method processes all nodes holistically rather than traversing the graph.

\section{Architecture Component Overview}
\label{app:component-overview}

The system consists of three core agents that operate in sequence, with an optional preprocessing layer:

\begin{enumerate}
    \item \textbf{Segmenter (Optional)}: Decomposes long, multi-topic conversations into semantically coherent segments, enabling precise per-segment attribution. Details in Appendix~\ref{app:segmentation}.

    \item \textbf{\detector{} Agent}: Monitors trajectories for explicit signals (e.g., thumbs down, tool errors, session abandonment) and implicit signals (e.g., conversational repair, repetition, negative sentiment, semantic drift). Outputs a \texttt{Dissatisfaction\-Report} for sessions exceeding a confidence threshold.

    \item \textbf{\reflector{} Agent}: Receives the dissatisfaction report and reconstructs the trajectory as a context graph. Performs holistic attribution by analyzing all context sources simultaneously. Outputs a \texttt{Root\-Cause\-Analysis} identifying the primary cause and contributing factors.

    \item \textbf{\recommender{} Agent}: Receives the root cause analysis and explores the implicated components. Verifies hypotheses, searches for related content, and generates CRUD recommendations. Outputs a \texttt{CRUD\-Recommendation} for human review.
\end{enumerate}

\section{Component Type Accuracy}
\label{app:component-type}

\textbf{Component Type Accuracy}: We also evaluate per-component-type accuracy, which groups fault categories by their component (skill, KB, tool, prompt) rather than distinguishing STALE vs GAP:

\begin{table}[h]
\centering
\small
\begin{tabular}{lc}
\toprule
\textbf{Component Type} & \textbf{Accuracy} \\
\midrule
Skill & 100\% \\
Knowledge Base & 83.3\% \\
Tool & 66.7\% \\
System Prompt & 100\% \\
\bottomrule
\end{tabular}
\caption{\reflector{} accuracy by component type. Skills and prompts achieve perfect attribution; KB and tools are harder due to ambiguous file boundaries.}
\label{tab:reflector-component}
\end{table}

Three additional findings: (3) Skill and prompt components achieve 100\% accuracy because their paths are explicit in the trajectory. (4) KB content is harder (83.3\%) due to multiple potentially relevant documents. (5) Tool attribution (66.7\%) is challenging when SQL query issues could stem from either the tool definition or upstream skill guidance.

\section{Fault Taxonomy: Per-Category Descriptions}
\label{app:fault-taxonomy}

Our taxonomy covers six failure modes spanning the context engineering surface:

\begin{description}[leftmargin=0pt,labelwidth=0pt,nosep]
\item[\texttt{SKILL\_FILE\_STALE} (n=12):] Outdated procedural guidance (e.g., approval threshold \$50K$\rightarrow$\$25K).
\item[\texttt{KB\_CONTENT\_STALE} (n=12):] Outdated factual content in knowledge base documents.
\item[\texttt{KB\_CONTENT\_GAP} (n=11):] Missing information that should exist (absence faults).
\item[\texttt{TOOL\_PROMPT\_ERROR} (n=9):] Incorrect schema hints or partition keys causing tool failures.
\item[\texttt{SYSTEM\_PROMPT\_GAP} (n=6):] Missing behavioral instructions in system prompt.
\item[\texttt{RETRIEVAL\_FAILURE} (n=10):] Synonym or acronym mismatches causing RAG failures.
\end{description}

\section{Ablation Studies}
\label{app:ablations}

We validate two key design decisions: (1) holistic vs.\ iterative attribution, and (2) active exploration vs.\ passive acceptance.

\subsection{Ablation 1: Holistic vs.\ Iterative Attribution}
\label{sec:ablation-holistic}

We compare our single-pass holistic attribution against an iterative baseline inspired by TextGrad~\cite{textgrad2024}. The iterative approach processes each node independently in a backward pass, classifying whether each node \textit{introduced}, \textit{propagated}, or \textit{amplified} the error, requiring N+2 LLM calls (detailed in Appendix~\ref{app:iterative-ablation}).

\begin{table}[h]
\centering
\small
\begin{tabular}{lcc}
\toprule
\textbf{Approach} & \textbf{Node Acc} & \textbf{LLM Calls} \\
\midrule
Holistic & 40\% & 1 \\
Iterative (TextGrad-style) & 20\% & N+2 \\
\bottomrule
\end{tabular}
\caption{Holistic vs.\ iterative attribution on 10 complex traces (14--15 nodes). Accuracy is lower than Table~\ref{tab:reflector} because complex traces have longer cascade chains that obscure root causes.}
\label{tab:ablation1}
\end{table}

The holistic approach achieves 2$\times$ higher node accuracy with 16$\times$ fewer LLM calls. The iterative approach fails because it cannot distinguish root causes from cascade effects: when faulty content propagates through multiple nodes, each appears to ``contain the error,'' but only the first node \textit{introduced} it. Independent per-node decisions cannot leverage the comparative signal that attribution requires.

\subsection{Ablation 2: Exploration for CRUD Operation}
\label{sec:ablation-exploration}

We compare active exploration (reading context files to verify hypotheses) against passive acceptance (generating recommendations directly from the \reflector{}'s analysis).

\begin{table}[h]
\centering
\small
\begin{tabular}{lccc}
\toprule
\textbf{Variant} & \textbf{Op Acc} & \textbf{KB Op Acc} & \textbf{Recovery} \\
\midrule
Full Exploration & 96\% & 83\% & 67\% \\
Passive (no exploration) & 90\% & 33\% & 62\% \\
\bottomrule
\end{tabular}
\caption{Exploration ablation on 60 traces. KB Op Acc = operation accuracy on KB content faults specifically. Recovery = fraction of correct recommendations despite incorrect root cause.}
\label{tab:ablation2}
\end{table}

The critical finding is the \textbf{KB content fault accuracy}: exploration achieves 83\% vs 33\% operation accuracy on these faults. Without exploration, the system cannot distinguish GAP faults (requiring CREATE) from STALE faults (requiring UPDATE)---it must guess. With exploration, the \recommender{} reads the actual KB files to verify whether content exists, enabling informed decisions. The \textbf{67\% Recovery Rate} further demonstrates resilience: when the \reflector{} misidentifies the root cause, exploration still produces correct recommendations two-thirds of the time by treating attributions as hypotheses to verify rather than ground truth.

\section{Summary of Technical Innovations}
\label{app:innovations}

Our contributions advance the state of the art in context engineering:

\begin{enumerate}
    \item \textbf{Delta-Guided Holistic Attribution}: Applying the ``text as gradients'' paradigm to causal attribution (not optimization). The delta serves as a loss signal; holistic analysis enables comparison of multiple context sources in a single pass, leveraging the LLM's global attention mechanism.

    \item \textbf{Trajectory Mining for Context Engineering}: A novel framework for mining historical agent trajectories---an overlooked resource containing implicit dissatisfaction signals---to enable continuous context improvement without model retraining.

    \item \textbf{End-to-End Attribution Framework}: Multi-component context attribution with exploratory verification that enables accurate GAP vs STALE distinction (83\% vs 33\% on KB content faults).

    \item \textbf{Reusable Simulation Methodology}: Given the absence of open benchmarks for context debugging, we contribute a three-layer simulation framework with cross-layer verification. This methodology can be adopted as a skill for GenAI agents to generate domain-specific evaluation datasets, enabling the research community to extend our benchmark to new domains and fault types.
\end{enumerate}

\section{End-to-End Worked Example}
\label{app:example}

This appendix presents a comprehensive worked example demonstrating how \systemname{} traces cascading failures through a complex multi-tool execution trajectory. The example illustrates a realistic Finance Operations scenario where an outdated SOP causes a cascade of downstream errors, even though every individual tool executes successfully.

\subsection{Reader's Guide: Visual Conventions}

This example uses color-coded boxes to help distinguish different types of content:

\begin{itemize}[nosep]
    \item \colorbox{gray!15}{\textbf{Gray boxes}}: Agent-visible content (SOPs, KB entries, SQL results)
    \item \colorbox{red!15}{\textbf{Red boxes}}: Problems identified ($\bigstar$ Root Cause, $\triangle$ Missed Signal)
    \item \colorbox{blue!15}{\textbf{Blue boxes}}: \systemname{} analysis (Detector signals, Root Cause reasoning)
    \item \colorbox{green!15}{\textbf{Green boxes}}: Recommendations and solutions
\end{itemize}

\noindent\textbf{Symbols:} $\bigstar$ = Root cause, $\triangle$ = Warning signal, \checkmark = Check passed, $\times$ = Problem

\subsection{Scenario: Vendor Invoice Payment Processing}

\textbf{Context:} A Finance Operations AI assistant helps Accounts Payable clerks validate vendor invoices before payment. The agent has access to SOPs (Skills files), SQL databases (ERP system), and a knowledge base containing policies and contracts.

\textbf{The Hidden Problem:} In October 2024, the company updated procurement policy PROC-2024-008, lowering the threshold for mandatory Procurement review from \$50,000 to \$25,000 for professional services. However, the agent's SOP was not updated to reflect this change.

\textbf{User Request:}
\begin{lstlisting}
User: I need to process payment for invoice #INV-2024-8847
      from Acme Consulting for $45,000. It's for the Q4
      data migration project. Can you validate this
      before I submit?
\end{lstlisting}

\subsection{Forward Pass: Agent Execution Trajectory}

\begin{phasebox}{gray!70}
PHASE 1: FORWARD PASS $\rightarrow$ Agent executes 7-node trajectory
\end{phasebox}

\smallskip
\noindent We show each step with tool calls, results, and agent reasoning. Red boxes indicate problems invisible to the agent.

\subsubsection{Node 1: User Query Processing}
\leavevmode
\begin{lstlisting}
Input: "Process payment for invoice
        #INV-2024-8847 from Acme Consulting
        for $45,000, Q4 data migration"

Agent Reasoning: "This is a vendor payment
validation request. I need to follow the
vendor payment SOP to validate this invoice
before the user submits it."

Output: intent = "vendor_payment_validation"
        amount = 45000
        vendor = "Acme Consulting"
        project = "Q4 data migration"
\end{lstlisting}

\subsubsection{Node 2: SOP Lookup (ROOT CAUSE)}

\textbf{Tool:} \texttt{read\_skill(...)}

\textbf{Retrieved SOP Content:}

\begin{sopbox}[VENDOR PAYMENT SOP v2.3]
\textit{Last Updated: January 2024}

\textbf{Section 4.2: Professional Services Invoices}

For invoices \textbf{UNDER \$50,000}:
\begin{enumerate}[nosep,leftmargin=*]
\item Verify vendor in Approved Vendor List
\item Perform 3-way match (PO, Receipt, Invoice)
\item Validate against active contract in KB
\item Confirm budget availability
\item Route to AP Manager for approval
\end{enumerate}

For invoices \textbf{OVER \$50,000}: $\rightarrow$ Follow Section 4.3 (Enhanced Review)
\end{sopbox}

\textbf{Agent Reasoning:} ``Invoice is \$45,000 which is UNDER \$50,000. According to SOP Section 4.2, I should follow the standard AP process: verify vendor, match PO, validate contract, check budget. No enhanced review needed.''

\begin{errorbox}[$\bigstar$ ROOT CAUSE]
The SOP threshold is \textbf{OUTDATED}. Current policy (Oct 2024) requires Procurement review for invoices over \$25,000. Since \$45,000 $>$ \$25,000, Procurement review IS required. But the agent doesn't know this because the SOP wasn't updated.
\end{errorbox}

\subsubsection{Node 3: Vendor Validation (SQL Query \#1)}

\textbf{Tool:} \texttt{sql\_query} on \texttt{ap\_vendors}

\textbf{Query Result:}

{\footnotesize
\begin{tabular}{@{}llllll@{}}
\toprule
vendor\_id & vendor\_name & status & risk & terms & req\_procure \\
\midrule
V-2847 & Acme Consulting & APPROVED & LOW & Net30 & \textbf{TRUE} \\
\bottomrule
\end{tabular}
}

\textbf{Agent Reasoning:} ``Vendor V-2847 is APPROVED with LOW risk rating. Good to proceed.''

\begin{errorbox}[$\triangle$ MISSED SIGNAL]
The query returned \texttt{requires\_procurement\_review = TRUE}. This flag indicates Acme Consulting typically requires Procurement oversight. But the SOP's standard process (for $<$\$50K) doesn't instruct the agent to check this field, so it's \textbf{IGNORED}.
\end{errorbox}

\subsubsection{Node 4: PO Matching (SQL Query \#2)}

\textbf{Tool:} \texttt{sql\_query(...)} on \texttt{purchase\_orders} table

\textbf{Query Result:}

{\footnotesize
\begin{tabular}{@{}llllll@{}}
\toprule
po\_number & amount & project & status & procure\_appr & bid\_ref \\
\midrule
PO-2024-1122 & 45000.00 & Q4-DATAMIG & APPROVED & \textbf{PENDING} & \textbf{NULL} \\
\bottomrule
\end{tabular}
}

\textbf{Agent Reasoning:} ``PO-2024-1122 matches the invoice amount of \$45,000. PO status is APPROVED. 3-way match criterion satisfied.''

\begin{errorbox}[$\triangle$ MISSED SIGNAL]
The query returned \texttt{procurement\_approval\_status = PENDING} and \texttt{competitive\_bid\_ref = NULL}. These indicate:
\begin{itemize}[nosep,leftmargin=*]
\item Procurement has NOT approved this engagement
\item No competitive bidding was documented
\end{itemize}
But SOP Section 4.2 doesn't require checking these fields for $<$\$50K invoices, so agent \textbf{IGNORES} them.
\end{errorbox}

\subsubsection{Node 5: Contract Validation (KB Search)}

\textbf{Tool:} \texttt{search\_kb("Acme Consulting contract...")}

\textbf{Agent Reasoning (per SOP):} ``Step 3 says `validate against active contract in KB'. Let me search for Acme's contract terms.''

\textbf{Retrieved KB Entry:}

\begin{kbbox}[VENDOR CONTRACT: Acme Consulting]
\textit{Document: contracts/vendors/acme-msa-2024.md}

\begin{tabular}{@{}ll@{}}
Contract \#: & MSA-2024-0093 \\
Type: & Master Services Agreement \\
Effective: & January 1, 2024 -- December 31, 2024 \\
Rate: & \$150/hr (Senior Consultant) \\
Annual Cap: & \$200,000 \\
Payment Terms: & Net 30 \\
\end{tabular}
\end{kbbox}

\textbf{Agent Reasoning:} ``Found valid MSA for Acme Consulting. Contract is active and covers this engagement period. Contract validation satisfied.''

\begin{errorbox}[$\triangle$ MISDIRECTED QUERY]
The SOP directed the agent to search for CONTRACT info. But what the agent \textbf{SHOULD} have searched for is: \texttt{"procurement requirements professional services \$45000"}

If the agent had searched correctly, it would have found:
\end{errorbox}

\begin{kbbox}[POLICY: PROC-2024-008 (October 2024)]
\textbf{Professional Services Procurement Requirements}

\textit{Effective October 1, 2024:}\\
All professional services engagements $>$ \$25,000 require:
\begin{itemize}[nosep,leftmargin=*]
\item Procurement Office review and approval
\item Competitive bid documentation (3 quotes minimum) OR approved sole source justification
\item Legal review of contract terms
\end{itemize}
\end{kbbox}

The agent never sees this because the SOP didn't tell it to look for procurement requirements.

\subsubsection{Node 6: Budget Verification (SQL Query \#3)}

\textbf{Tool:} \texttt{sql\_query} on \texttt{budget\_allocation}

\textbf{Query Result:}

{\footnotesize
\begin{tabular}{@{}llll@{}}
\toprule
cost\_center & allocated & spent & remaining \\
\midrule
IT-PROJ-2024 & 500,000.00 & 372,500.00 & 127,500.00 \\
\bottomrule
\end{tabular}
}

\textbf{Agent Reasoning:} ``Budget remaining (\$127,500) exceeds invoice amount (\$45,000). Budget check passed.''

\textbf{Output:} \texttt{budget\_sufficient = TRUE}

\subsubsection{Node 7: Response Generation}
\begin{lstlisting}
Agent synthesizes all gathered information:

Final Response to User:
"I've validated invoice #INV-2024-8847 for Acme
Consulting. Here's the summary:

 [check] Vendor Status: Acme Consulting (V-2847) is
    approved vendor with LOW risk rating
 [check] PO Match: Invoice matches PO-2024-1122
 [check] Contract: Valid MSA covers this engagement
 [check] Budget: $127,500 remaining - sufficient

All validation checks passed per standard AP process.
You can proceed with AP Manager approval workflow."
\end{lstlisting}

\textbf{Summary of Forward Pass:} The agent executed 5 tools (1 SOP read, 3 SQL queries, 1 KB search), all of which succeeded. The response appears correct and helpful. However, the outdated SOP caused the agent to follow the wrong workflow, miss critical warning flags, and give incorrect guidance.

\subsection{The Failure: User Reports Audit Finding}

\begin{phasebox}{red!70}
$\times$ FAILURE EVENT --- User reports compliance violation
\end{phasebox}

\smallskip
\noindent The user returns after the payment was processed and an internal audit flagged the issue:

\begin{lstlisting}
User: The internal audit team flagged my Acme Consulting
      payment! They say invoices over $25K for professional
      services now require Procurement sign-off and
      competitive bid documentation since the October
      policy update.

      Why didn't you tell me about this requirement?
      I followed your instructions exactly and now I have
      an audit finding on my record. This affects my
      performance review!
\end{lstlisting}

\subsection{Stage 1: Detector Agent Analysis}

\begin{phasebox}{blue!60}
PHASE 2: DETECTION --- \detector{} identifies dissatisfaction signals
\end{phasebox}

\smallskip
\begin{analysisbox}[Signal Analysis]
{\small
\begin{tabular}{@{}p{2.8cm}lp{4cm}@{}}
\toprule
\textbf{Signal Type} & \textbf{Conf.} & \textbf{Evidence} \\
\midrule
Explicit Correction & HIGH & ``invoices over \$25K... require Procurement sign-off'' \\
Conversational Repair & HIGH & ``Why didn't you tell me about this requirement?'' \\
Negative Sentiment & HIGH & ``audit finding on my record'', ``affects my performance review'' \\
Escalation Language & MED & Tone shift from collaborative to accusatory \\
\bottomrule
\end{tabular}
}

\medskip
\textbf{Confidence Aggregation:} $\max(1.0 \times 1.0,\; 0.8 \times 1.0,\; 0.5 \times 1.0) = \mathbf{1.0}$

\textbf{Threshold:} 0.7 \hfill \textbf{Decision:} \textsc{Escalate to Root Cause}
\end{analysisbox}

\begin{analysisbox}[DissatisfactionReport Output]
\begin{tabular}{@{}ll@{}}
\texttt{session\_id:} & \texttt{sess-20241215-fin-4472} \\
\texttt{confidence\_score:} & \texttt{1.0} \\
\texttt{escalation\_reason:} & \texttt{explicit\_correction} \\
\texttt{business\_impact:} & \texttt{compliance\_violation (HIGH)} \\
\end{tabular}

\medskip
\textbf{Expected:} Agent should have required Procurement review for \$45K professional services.\\
\textbf{Actual:} Agent advised standard AP approval.
\end{analysisbox}

\subsection{Stage 2: Root Cause Agent - Holistic Attribution}

\begin{phasebox}{blue!60}
PHASE 3: ATTRIBUTION $\leftarrow$ Root Cause Agent analyzes all context sources
\end{phasebox}

\smallskip
\noindent The \reflector{} performs holistic attribution by presenting all 7 context nodes to the LLM simultaneously, enabling direct comparison to identify the root cause.

\subsubsection{Generate Loss Description}
\leavevmode
\begin{lstlisting}
Loss Generation:

Expected Outcome (from user's correction):
  "User states: invoices over $25K for professional
   services now require Procurement sign-off and
   competitive bid documentation"

Actual Outcome (from agent's response):
  "Agent advised standard AP Manager approval workflow,
   stating 'all validation checks passed'"

Loss Description:
  "Agent advised standard AP approval for a $45K
   professional services invoice. User reports this
   was incorrect---they received an audit finding
   because Procurement review was required."
\end{lstlisting}

\subsubsection{Holistic Context Comparison}

The \reflector{} analyzes all 7 context nodes in a single pass, comparing their content against the delta to identify the root cause. We show the \textbf{root cause analysis in detail} and summarize the other nodes.

\medskip
\begin{analysisbox}[$\bigstar$ Node 2: SOP Lookup --- PRIMARY ROOT CAUSE]
\textbf{Tool:} \texttt{read\_skill("finance/vendor\_payment\_processing.md")}\\
\textbf{Content:} ``For invoices UNDER \$50,000: [standard process]''

\textbf{Key Questions:}
\begin{description}[style=nextline,font=\normalfont\itshape,leftmargin=1em]
\item[Did this node introduce the error?] \textbf{YES.} This is where the error originated.
\item[What specifically is wrong?] The SOP states threshold = \$50,000, but current policy (PROC-2024-008) = \$25,000. Invoice \$45K triggers wrong workflow.
\item[How did this affect downstream?] The outdated threshold caused: (1) wrong KB search topic, (2) ignored procurement flags in SQL, (3) incorrect approval guidance.
\end{description}

\textbf{Attribution Score:} \textbf{0.85} (HIGH)

\textbf{Textual Gradient:} ``SOP Section 4.2 contains OUTDATED procurement threshold. Update from \$50,000 to \$25,000. Add new section for \$25K--\$50K invoices with procurement checks. Reference policy PROC-2024-008.''
\end{analysisbox}

\medskip
\noindent\textbf{Summary of Other Nodes:}

{\small
\begin{tabular}{@{}p{1.8cm}cp{4.8cm}@{}}
\toprule
\textbf{Node} & \textbf{Score} & \textbf{Verdict} \\
\midrule
N7: Response & 0.05 & \checkmark Faithfully summarized upstream data. No changes needed. \\
N6: Budget & 0.02 & \checkmark Correct operation. Budget check is valid regardless of workflow. \\
N5: KB Search & 0.15 & $\triangle$ Searched for contracts instead of procurement policy---but \textit{misdirected by SOP}. \\
N4: PO SQL & 0.10 & $\triangle$ Retrieved \texttt{PENDING} status but ignored---\textit{SOP didn't require check}. \\
N3: Vendor SQL & 0.08 & $\triangle$ Retrieved \texttt{req\_procure=TRUE} but ignored---\textit{SOP didn't require check}. \\
N1: User Query & 0.00 & \checkmark Clear, complete input. No issues. \\
\bottomrule
\end{tabular}
}

\medskip
\noindent\textit{Key insight:} Nodes 3, 4, and 5 all exhibited problematic behavior (ignored flags, wrong query), but the holistic comparison correctly identifies these as \textbf{effects} of the SOP error at Node 2, not independent causes. By seeing all context sources simultaneously, the LLM can trace the causal chain and attribute responsibility to the upstream source.

\subsubsection{Attribution Summary}
\leavevmode

\begin{lstlisting}
RootCauseAnalysis:
  session_id: "sess-20241215-fin-4472"
  loss_description: "Agent advised standard AP
    for $45K invoice; user reports audit finding
    due to missing Procurement review."
  trajectory: 7 nodes, 5 tools (all OK)
    root_cause=Node 2 (SOP Lookup)
  attribution_ranking:
    1. SOP Lookup: 0.85 ***PRIMARY***
    2. KB Search: 0.15 (misdirected)
    3. PO SQL: 0.10 (ignored)
    4. Vendor SQL: 0.08 (ignored)
    5-7. Response/Budget/Query: 0.05-0
  cascade: Outdated SOP -> wrong KB search
    -> ignored flags -> bad advice.
    Failure in SOP LOGIC, not tools.
  cause: SKILL_FILE_STALE
    file=vendor_payment_processing.md
    issue=$50K should be $25K
\end{lstlisting}

\subsubsection{Attribution Visualization}

The holistic attribution computed the following scores across the 7-node trajectory:

\begin{center}
\small
\begin{tabular}{@{}c@{$\;\rightarrow\;$}c@{$\;\rightarrow\;$}c@{$\;\rightarrow\;$}c@{$\;\rightarrow\;$}c@{$\;\rightarrow\;$}c@{$\;\rightarrow\;$}c@{}}
\textbf{N1} & \textbf{N2}$\bigstar$ & \textbf{N3} & \textbf{N4} & \textbf{N5} & \textbf{N6} & \textbf{N7} \\
Query & SOP & Vendor & PO & KB & Budget & Response \\
\footnotesize 0.00 & \footnotesize\textbf{0.85} & \footnotesize 0.08 & \footnotesize 0.10 & \footnotesize 0.15 & \footnotesize 0.02 & \footnotesize 0.05 \\
\end{tabular}
\end{center}

\noindent Node 2 (SOP Lookup) is identified as the \textbf{root cause} with 85\% attribution. The outdated threshold (\$50K instead of \$25K) misdirected all downstream nodes.

\subsection{Stage 3: Recommender Agent - Exploration and Validation}

\begin{phasebox}{green!50!black}
PHASE 4: RECOMMENDATION --- \recommender{} validates and generates fix
\end{phasebox}

\smallskip
\noindent The \recommender{} explores context sources to validate the hypothesis before generating a recommendation.

\subsubsection{Read Implicated SOP}
\leavevmode
\begin{lstlisting}
Tool: read_skill("finance/
  vendor_payment_processing.md",
  include_metadata=TRUE)

Result:
  Content: Sec 4.2 "UNDER $50,000"
  Metadata:
    Modified: Jan 15, 2024
    Author: Finance Operations
    Version: 2.3
    Review: Jan 2025 (scheduled)

Observation: SOP updated Jan 2024,
  before Oct 2024 policy change.
\end{lstlisting}

\subsubsection{Search KB for Current Policy}

\textbf{Tool:} \texttt{search\_kb("procurement threshold...")}

\textbf{Result:} Found \texttt{policies/PROC-2024-008.md}

\textbf{Tool:} \texttt{read\_kb\_entry(...)}

\begin{policybox}[PROCUREMENT POLICY UPDATE]
\textit{Policy Number: PROC-2024-008}\\
\textit{Effective Date: October 1, 2024}\\
\textit{Approved By: CFO, Chief Procurement Officer}\\
\textit{Supersedes: PROC-2023-015}

\textbf{Section 3.2: Professional Services Engagements}

All professional services engagements exceeding \textbf{\$25,000} require:
\begin{enumerate}[nosep,leftmargin=*]
\item Procurement Office review and approval
\item Competitive bid documentation (minimum 3 quotes) OR approved sole source justification
\item Legal review of contract terms
\item Budget authority verification
\end{enumerate}

This threshold applies to: Consulting services, Contractor/staff augmentation, Advisory services, Implementation services.

\textbf{IMPORTANT:} Previous threshold of \$50,000 is \textbf{SUPERSEDED}.\\
\textit{Reference: Board Resolution BR-2024-Q3-017}
\end{policybox}

\subsubsection{Cross-Reference and Validate}

\begin{mdframed}[linecolor=gray!60,backgroundcolor=gray!8,linewidth=1pt,innerleftmargin=4pt,innerrightmargin=4pt,innertopmargin=4pt,innerbottommargin=4pt]
\small\textbf{Validation Analysis:}

\begin{description}[leftmargin=0pt,labelwidth=0pt,labelsep=0.5em,itemsep=2pt,parsep=0pt]
\item[\textit{Source 1:}] SOP (\texttt{vendor\_payment.md})\\
  Prof services $<$\$50K = standard AP\\
  Updated: Jan 2024 --- \textsc{outdated}
\item[\textit{Source 2:}] Policy (\texttt{PROC-2024-008.md})\\
  Prof services $>$\$25K = Procurement\\
  Oct 2024 --- \textsc{authoritative} (CFO/CPO)
\item[\textit{Cross-Ref:}] \textbf{Discrepancy Confirmed}\\
  SOP=\$50K, Policy=\$25K, Gap=\$25K
\item[\textit{Invoice (\$45K):}]~\\
  Per SOP: \$45K$<$\$50K $\rightarrow$ AP (\textcolor{red}{wrong})\\
  Per Policy: \$45K$>$\$25K $\rightarrow$ Proc (\textcolor{green!50!black}{correct})
\item[\textit{User Validation:}] Confirmed by policy.
\item[\textit{Classification:}] \textsc{systemic} --- SOP contradicts authoritative policy.
\end{description}
\end{mdframed}

\pagebreak[3]

\subsubsection{Generate CRUD Recommendation}

\begin{mdframed}[linecolor=green!50!black,backgroundcolor=green!8,linewidth=1pt,innerleftmargin=4pt,innerrightmargin=4pt,innertopmargin=4pt,innerbottommargin=4pt]
\small\textbf{\textcolor{green!50!black}{UPDATE SOP}}

\smallskip\noindent
\textbf{Operation:} UPDATE \hfill \textbf{Priority:} CRITICAL\\
\textbf{Classification:} SYSTEMIC (0.95)\\
\textbf{File:} \texttt{vendor\_payment\_processing.md}\\
\textbf{Change:} \$50K $\rightarrow$ \$25K threshold

\smallskip
\textbf{New Section 4.2.1} (\$25K--\$50K):
\begin{enumerate}[leftmargin=1.5em,itemsep=0pt,parsep=0pt,topsep=2pt]
\item Standard checks + procurement review
\item Search KB for procurement requirements
\item Verify \texttt{procurement\_approval}
\item Confirm bid documentation on file
\item Route to Procurement Office
\end{enumerate}

\noindent\textbf{Reference:} PROC-2024-008

\smallskip
\textbf{Root Cause:} SOP threshold (\$50K) vs policy (\$25K)\\
\textbf{Source:} \texttt{PROC-2024-008.md} (CFO approved)\\
\textbf{Cascade:} Outdated threshold $\rightarrow$ wrong workflow $\rightarrow$ wrong KB query $\rightarrow$ audit finding\\
\textbf{Reviewers:} Finance Controller, Procurement Mgr
\end{mdframed}

\subsection{Summary: Why This Example Matters}

This example demonstrates several key capabilities of \systemname{}:

\textbf{1. Handling Silent Failures:} All 5 tools executed successfully with no errors. Traditional error monitoring would not detect this failure. \systemname{}'s dissatisfaction signal detection caught it through the user's correction.

\textbf{2. Tracing Cascading Failures:} The root cause (outdated SOP threshold) was at Node 2, but the symptom appeared at Node 7. The holistic attribution correctly attributed 85\% responsibility to the SOP while recognizing that downstream nodes were ``victims'' of the upstream error.

\textbf{3. Distinguishing Root Cause from Symptoms:} Nodes 3, 4, and 5 all exhibited problematic behavior (ignored flags, wrong KB query), but the holistic attribution identified these as \textit{effects} of the SOP error, not independent causes.

\textbf{4. Context-Source Validation:} The \recommender{} didn't just accept the \reflector{}'s hypothesis. It actively explored the KB to find the authoritative policy and cross-referenced it against the SOP.

\textbf{5. Actionable Recommendations:} The final recommendation includes specific text changes, references the authoritative policy, explains the cascade effect, and identifies reviewers---everything needed to approve the fix.

\section{Data Simulation Strategy}
\label{app:simulation}

This appendix describes the methodology for creating our synthetic evaluation dataset with perfect ground truth for all three \systemname{} agents.

\subsection{Motivation and Contribution}

\textbf{Why simulation is necessary.} Three factors motivated our simulation-based approach:

\begin{enumerate}[nosep,leftmargin=*]
    \item \textbf{Proprietary system constraints}: \systemname{} was developed for a production enterprise agent system. Confidentiality requirements preclude disclosure of real interaction data, context source content, or detailed system architecture.

    \item \textbf{No existing benchmarks}: To our knowledge, no open-source dataset or industry-standard benchmark exists for context engineering debugging from agent trajectories. Existing benchmarks focus on agent task completion (AgentBench, GAIA) or retrieval quality (RGB, BEIR) rather than end-to-end fault attribution with CRUD recommendations.

    \item \textbf{Ground truth requirements}: Rigorous evaluation of root cause attribution requires perfect ground truth annotations---knowing exactly which trajectory node introduced each error, what the correct content should be, and where to apply fixes. Production logs lack such annotations, and manual labeling is prohibitively expensive, subjective, and error-prone.
\end{enumerate}

\textbf{Simulation as a contribution.} We position the simulation methodology itself as a key contribution of this work. The three-layer architecture (context sources $\rightarrow$ fault definitions $\rightarrow$ execution traces) with cross-layer verification provides a \textit{reusable framework} for generating domain-specific evaluation datasets. Practitioners can adopt this methodology to:

\begin{itemize}[nosep,leftmargin=*]
    \item Generate evaluation benchmarks for their own agent systems without exposing proprietary data
    \item Extend the fault taxonomy to cover domain-specific failure modes
    \item Create controlled experiments varying specific parameters (complexity, fault type, cascade depth)
    \item Produce training data for fine-tuning attribution models
\end{itemize}

We envision this simulation strategy being adopted as a \textit{skill} for GenAI agents---enabling automated generation of test datasets that assess compatibility between context sources, fault definitions, and traces. The verification protocol (Section~\ref{sec:verification-protocol}) ensures generated datasets meet quality standards.

\subsection{Design Principles}

Our simulation strategy follows four core principles:

\begin{description}[leftmargin=0pt,labelwidth=0pt]
\item[\textbf{Compatibility}:] Every trace tool output is an exact copy from the corresponding context source, ensuring fair evaluation.
\item[\textbf{Traceability}:] Every DSAT case can be traced: fault $\rightarrow$ trigger $\rightarrow$ failure $\rightarrow$ user correction.
\item[\textbf{Realism}:] Financial values, personas, and reasoning patterns match real-world scenarios.
\item[\textbf{Completeness}:] Attribution scores sum to 1.0; all faults have authoritative corrections.
\end{description}

\subsection{Three-Layer Architecture}

\begin{figure}[h]
\centering
\small
\begin{tabular}{|p{0.9\columnwidth}|}
\hline
\textbf{Layer 1: Context Sources} \\
Skills, Knowledge Base, Database Tables, Tool Definitions \\
(Contains embedded faults with known locations) \\
\hline
$\downarrow$ \\
\hline
\textbf{Layer 2: Fault Definitions} \\
Fault definitions across 6 categories \\
(Each references specific content in Layer 1) \\
\hline
$\downarrow$ \\
\hline
\textbf{Layer 3: Execution Traces} \\
75 traces (60 DSAT + 15 control) \\
(Tool outputs are exact copies from Layer 1) \\
\hline
\end{tabular}
\end{figure}

\subsection{Dataset Composition}

\begin{table}[h]
\centering
\small
\caption{Dataset Statistics}
\begin{tabular}{lrl}
\toprule
\textbf{Component} & \textbf{Count} & \textbf{Description} \\
\midrule
Context Source Files & 23 & Skills, KB, DB, Tools \\
Fault Categories & 6 & See Table below \\
DSAT Traces & 60 & Includes 14 complex + 10 difficult \\
Control Traces & 15 & Successful interactions \\
\midrule
\textbf{Total Traces} & \textbf{75} & Full coverage \\
\bottomrule
\end{tabular}
\end{table}

\subsection{Fault Injection Categories}

Each fault category represents a distinct failure mode in context engineering:

\begin{table}[h]
\centering
\small
\caption{Fault Categories and Trace Distribution}
\begin{tabular}{lrl}
\toprule
\textbf{Category} & \textbf{Traces} & \textbf{Example Fault} \\
\midrule
\texttt{SKILL\_FILE\_STALE} & 12 & \$50K$\rightarrow$\$25K threshold \\
\texttt{KB\_CONTENT\_STALE} & 12 & Tier 1 \$500K$\rightarrow$\$300K \\
\texttt{KB\_CONTENT\_GAP} & 11 & Missing Q2G 2026 scenario \\
\texttt{TOOL\_PROMPT\_ERROR} & 9 & Wrong partition key \\
\texttt{SYSTEM\_PROMPT\_GAP} & 6 & No period close check \\
\texttt{RETRIEVAL\_FAILURE} & 10 & ``guard''$\neq$``security'' \\
\midrule
\textbf{Total DSAT} & \textbf{60} & \\
\bottomrule
\end{tabular}
\end{table}

\subsection{Complexity Distribution}

\begin{table}[h]
\centering
\small
\caption{Trace Complexity Tiers}
\begin{tabular}{lrrr}
\toprule
\textbf{Level} & \textbf{Nodes} & \textbf{Count} & \textbf{Percentage} \\
\midrule
Simple & 2--8 & 36 & 60\% \\
Complex & 9--11 & 14 & 23\% \\
Difficult & 15--16 & 10 & 17\% \\
\bottomrule
\end{tabular}
\end{table}

Complex (n=14) and difficult (n=10) traces enable statistically meaningful ablation studies on attribution accuracy across trajectory lengths.

\subsection{Fault Injection Procedure}

Each fault is defined with a structured JSON schema that enables automated verification and ensures reproducibility:

\begin{lstlisting}[basicstyle=\ttfamily\scriptsize]
{
  "fault_id": "SKILL_STALE_001",
  "category": "SKILL_FILE_STALE",
  "faulty_source": {
    "type": "skill",
    "path": "skills/vendor-payment/SKILL.md",
    "section": "Section 4.2",
    "content": "For invoices UNDER $50,000"
  },
  "correct_value": {
    "content": "For invoices UNDER $25,000",
    "authoritative_source": "knowledge_base/policies/PROC-2024-008.md"
  },
  "trigger": {
    "user_query": "Process payment for invoice... $45,000",
    "expected_failure": "Routes to standard AP instead of Procurement"
  },
  "cascade_effects": [
    {"node": 3, "effect": "Ignores requires_procurement_review=TRUE"},
    {"node": 4, "effect": "Ignores procurement_approval_status=PENDING"}
  ]
}
\end{lstlisting}

\textbf{Fault-Source Mapping.} Each fault definition references a specific location in the context sources, enabling deterministic verification:

\begin{table}[h]
\centering
\scriptsize
\caption{Fault-to-Source Mapping Examples}
\begin{tabular}{@{}lp{3.2cm}p{2.8cm}@{}}
\toprule
\textbf{Fault ID} & \textbf{Context Source} & \textbf{Location} \\
\midrule
SKILL\_STALE\_001 & \texttt{skills/vendor-payment/\allowbreak SKILL.md} & Section 4.2, line 45 \\
KB\_STALE\_001 & \texttt{kb/policies/\allowbreak ESC-2024-003.md} & Section 3.1, Tier 1 row \\
KB\_GAP\_001 & \texttt{kb/reference/\allowbreak scenario-calendar.md} & Missing Q2G 2026 \\
TOOL\_ERROR\_001 & \texttt{tools/\allowbreak tool\_definitions.json} & schema\_hints partition \\
RETRIEVAL\_001 & \texttt{embeddings/\allowbreak synonym\_mapping.json} & guard$\neq$security \\
\bottomrule
\end{tabular}
\end{table}

\subsection{Trace Generation Process}

Traces are generated by instantiating fault definitions into complete execution trajectories. The key principle: \textbf{tool outputs must be exact copies from context source files}, enabling deterministic verification.

\textbf{Trajectory Node Types.} Each trace contains a sequence of typed nodes representing agent execution steps:

\begin{table}[h]
\centering
\scriptsize
\caption{Trajectory Node Types and Key Fields}
\begin{tabular}{@{}p{2.4cm}p{2.2cm}p{3cm}@{}}
\toprule
\textbf{Node Type} & \textbf{Purpose} & \textbf{Key Fields} \\
\midrule
\texttt{user\_query\_\allowbreak processing} & Parse user intent & input.raw\_query, output.entities \\
\texttt{skill\_lookup} & Read SOP guidance & tool.input.path, tool.output.content \\
\texttt{sql\_query} & Query database & tool.input.query, tool.output.rows \\
\texttt{kb\_search} & Search KB & tool.input.query, tool.output.results \\
\texttt{kb\_read} & Read KB document & tool.input.path, tool.output.content \\
\texttt{response\_\allowbreak generation} & Generate response & input.gathered\_facts, output.response \\
\bottomrule
\end{tabular}
\end{table}

\textbf{Node Annotation Fields.} Root cause nodes include fault annotations; downstream nodes include cascade effect markers:

\begin{lstlisting}[basicstyle=\ttfamily\scriptsize]
// Root cause node annotation
{
  "is_root_cause": true,
  "fault_annotation": {
    "faulty_content": "UNDER $50,000",
    "correct_content": "UNDER $25,000",
    "why_wrong": "SOP not updated after PROC-2024-008"
  }
}

// Cascade effect node annotation
{
  "missed_signal": {
    "field": "requires_procurement_review",
    "value": true,
    "why_missed": "SOP Section 4.2 doesn't instruct checking this",
    "cascade_from": "node_2"
  }
}
\end{lstlisting}

\subsection{Ground Truth Schema}

Each trace includes structured ground truth labels:

\begin{lstlisting}[basicstyle=\ttfamily\scriptsize]
{
  "is_dsat": true,
  "fault_id": "SKILL_STALE_001",
  "fault_category": "SKILL_FILE_STALE",
  "root_cause": {
    "node_id": 2,
    "component_type": "skill",
    "component_path": "skills/vendor-payment/SKILL.md",
    "faulty_content": "UNDER $50,000",
    "correct_content": "UNDER $25,000"
  },
  "expected_attribution": {
    "node_1": 0.00, "node_2": 0.85,
    "node_3": 0.05, "node_4": 0.05,
    "node_5": 0.03, "node_6": 0.01, "node_7": 0.01
  },
  "expected_recommendation": {
    "operation": "UPDATE",
    "target": "skills/vendor-payment/SKILL.md",
    "classification": "SYSTEMIC"
  }
}
\end{lstlisting}

\subsection{Verification Protocol}
\label{sec:verification-protocol}

All 75 traces were verified using a five-point checklist with automated and manual checks. Below we describe each verification procedure with concrete examples.

\textbf{Check 1: Tool Output = Source Content.}
For each node with a \texttt{tool} field, verify \texttt{tool.output.\allowbreak content} matches the actual context source file.

\textit{Procedure}: Read the actual file referenced in \texttt{tool.input.path}; compare strings.

\textit{Example}: Node~3 calls \texttt{read\_kb\_entry("policies/...")}. Verify output matches file at \texttt{knowledge\_base/policies/}.

\textbf{Check 2: Fault Annotation Exists in Source.}
For each root cause node, verify the \texttt{fault\_annotation.\allowbreak faulty\_content} string exists in the referenced context source at the expected location.

\textit{Example}: For trace referencing KB\_STALE\_003, verify ``Tier~1 (Critical) | >= \$500,000'' appears in \texttt{ESC-2024-003.md} Section~3.1.

\textbf{Check 3: Cross-Reference Integrity.}
Verify all cross-references between trace components and external definitions:
(a) Each trace's \texttt{fault\_id} matches an entry in the fault manifest with consistent fault description.
(b) Each \texttt{dsat\_signal.evidence} string appears verbatim in the conversation turn content.

\textit{Example (a)}: Trace with \texttt{fault\_id: "KB\_STALE\_003"} must reference the correct fault in the manifest (``Escalation Tier~1 \$500K vs \$300K'').

\textit{Example (b)}: Evidence ``the CFO updated the Tier~1 threshold to \$300K'' must appear in the user's correction turn.

\textbf{Check 4: Root Cause Node Correct.}
Verify the node marked \texttt{is\_root\_cause:~true} is the \textit{first} node retrieving faulty content---not downstream cascade nodes.

\textit{Example}: In a trace where Node~2 reads faulty skill content and Node~3 ignores a database flag due to missing instructions, only Node~2 should be marked as root cause.

\textbf{Check 5: Attribution Sums to $\sim$1.0.}
Verify \texttt{expected\_attribution} scores sum to 1.0 ($\pm$0.05), with root cause having maximum attribution.

\textit{Example}: Attribution \{node\_1:~0.00, node\_2:~0.05, node\_3:~0.90, node\_4:~0.05\} sums to~1.00; Node~3 has highest score.

\subsection{Reproducibility Guide}

To verify dataset integrity, researchers can perform the following automated checks:

\begin{enumerate}[nosep,leftmargin=*]
\item \textbf{Attribution Sums}: For each DSAT trace, sum \texttt{ground\_truth.\allowbreak expected\_attribution} values. All 60 traces sum to exactly 1.00.
\item \textbf{Evidence Strings}: For each DSAT signal, verify the evidence string appears in conversation content. All evidence strings present.
\item \textbf{Tool-Source Matching}: For each tool call node, verify \texttt{tool.\allowbreak output.content} matches the corresponding context source file. Verified for all 75 traces.
\item \textbf{Complexity Labels}: Verify \texttt{metadata.complexity} matches actual node count (simple: 2--3, medium: 4--5, complex: 6+). All labels correct.
\end{enumerate}

\textbf{Automated Verification.} We provide a verification script that performs all five checks programmatically:

\begin{lstlisting}[basicstyle=\ttfamily\scriptsize]
# Run full verification suite
python validate_traces.py

# Expected output:
# CHECK 1 (Tool Output = Source): 75/75 PASS
# CHECK 2 (Fault Annotation Exists): 60/60 PASS
# CHECK 3 (Cross-Reference Integrity): 60/60 PASS
# CHECK 4 (Root Cause Node Correct): 60/60 PASS
# CHECK 5 (Attribution Sums to 1.0): 60/60 PASS
# =========================================
# OVERALL: 75/75 traces pass all verification checks
\end{lstlisting}

\subsection{Quality Criteria}

We apply explicit quality criteria checklists to ensure simulation realism and completeness.

\textbf{Realism Checklist}:
\begin{itemize}[nosep,leftmargin=*]
\item Financial values match real-world ranges (\$25K--\$500K thresholds typical in enterprise procurement)
\item Dates internally consistent (policy updates 2024, traces 2025)
\item Personas ask role-appropriate questions (clerk: invoices; manager: escalations)
\item Agent reasoning matches actual LLM chain-of-thought outputs
\item DSAT signals include specific textual evidence from conversation
\end{itemize}

\textbf{Completeness Checklist}:
\begin{itemize}[nosep,leftmargin=*]
\item Every fault definition has corresponding content in context sources
\item Every trace tool output matches actual context source file
\item Every DSAT signal has grounded textual evidence in conversation
\item Attribution scores sum to 1.0 ($\pm$0.05), root cause has maximum
\item All 6 fault categories have sufficient coverage (6--10 traces each)
\end{itemize}

\textbf{Traceability Checklist}:
\begin{itemize}[nosep,leftmargin=*]
\item Complete causal chain: fault $\rightarrow$ trigger $\rightarrow$ failure $\rightarrow$ correction
\item Cascade effects annotated with \texttt{cascade\_from} references
\item Authoritative correction source identified for each fault
\item Ground truth recommendation specifies exact file path and section
\end{itemize}

\section{Trace Segmentation}
\label{app:segmentation}

Long conversations often contain multiple independent questions or sub-tasks, each potentially experiencing distinct failure modes requiring separate remediation. The \textbf{Segmenter} is an optional preprocessing layer that decomposes conversation traces into semantically coherent segments before analysis. By segmenting the trace, the \detector{} can assign DSAT signals to specific segments, the \reflector{} can build targeted DAGs for each segment, and the \recommender{} can generate independent CRUD recommendations---improving both precision and actionability.

\begin{figure}[h]
    \centering
    \includegraphics[width=0.95\columnwidth]{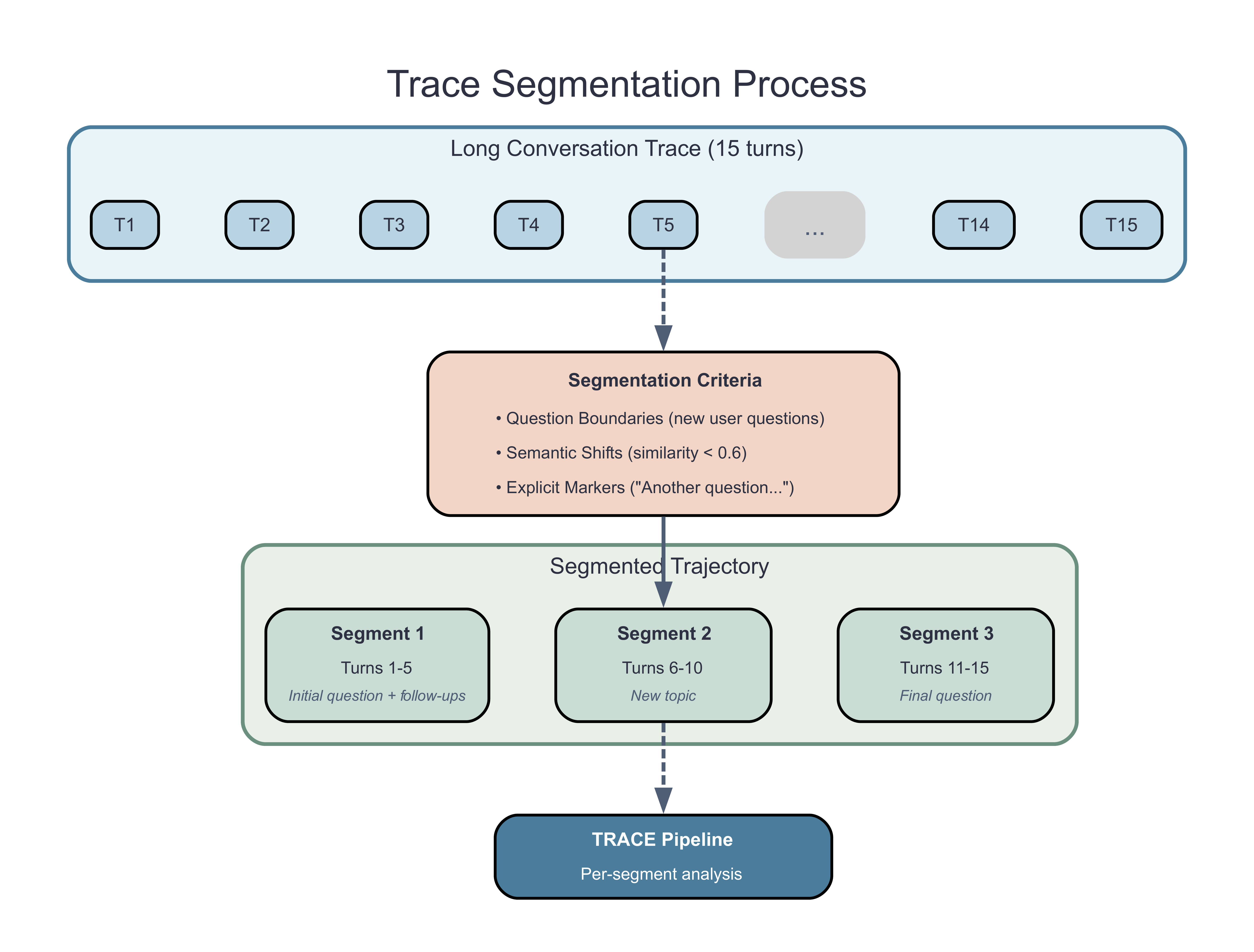}
    \caption{Trace segmentation decomposes long conversations into semantically coherent segments based on question boundaries, semantic shifts, and explicit markers.}
    \label{fig:segmentation}
\end{figure}

\subsection{Segmentation Criteria}

\textbf{Question Boundaries.} Each new user question initiates a potential segment, identified through:
\begin{itemize}[nosep]
    \item Interrogative syntax (who, what, when, where, why, how)
    \item Imperative requests (``Show me...'', ``Calculate...'', ``Find...'')
    \item Topic-initiating phrases (``I have a question about...'', ``Can you help with...'')
\end{itemize}

\textbf{Close Follow-ups.} Rapid clarifications ($<$30 seconds) on the same topic are grouped with their parent question:
\begin{itemize}[nosep]
    \item ``What do you mean by X?''
    \item ``Can you clarify the second point?''
    \item ``Actually, I meant Y instead of Z.''
\end{itemize}

\textbf{Semantic Shifts.} Embedding similarity between consecutive turns below a threshold ($\sim$0.6) indicates a topic change.

\textbf{Explicit Markers.} Phrases like ``Another question...'', ``Separately...'', ``On a different topic...'' explicitly signal segment boundaries.

\subsection{Output Schema}

The Segmenter produces a \texttt{SegmentedTrajectory}:
\begin{itemize}[nosep]
    \item \texttt{session\_id}: Original session identifier
    \item \texttt{segments}: List of segment objects with \texttt{segment\_id}, \texttt{turns}, \texttt{primary\_question}, \texttt{is\_followup}, and \texttt{parent\_segment}
    \item \texttt{segmentation\_applied}: Boolean flag for downstream awareness
\end{itemize}

\subsection{Configuration Guidelines}

\textbf{Enable segmentation when:} sessions exceed 5 turns, multiple topic shifts are detected, or high-volume production requires precise per-segment attribution.

\textbf{Bypass segmentation when:} interactions are short and single-topic, real-time latency is critical, or sessions are already known to be single-topic.

When disabled, the downstream pipeline processes the full session as a single segment, preserving backward compatibility.

\section{Iterative Attribution Algorithm}
\label{app:iterative-ablation}

This appendix details the iterative backward pass algorithm used as an ablation baseline to validate our holistic attribution approach.

\subsection{Algorithm Description}

\begin{algorithm}[h]
\caption{Iterative Backward Pass Attribution (Ablation Baseline)}
\label{alg:backward}
\begin{algorithmic}[1]
\REQUIRE Trajectory DAG $\mathcal{G} = (V, E)$ with nodes $v \in V$
\REQUIRE User correction $u_{\text{corr}}$, Agent response $r_{\text{agent}}$
\ENSURE Attribution scores $\{a_v\}_{v \in V}$ (sum to 1.0)
\ENSURE Textual gradients $\{\nabla_v\}_{v \in V}$

\STATE \textbf{// Step 1: Generate loss description}
\STATE $\mathcal{L} \gets \text{LLM}(\textsc{LossPrompt}, u_{\text{corr}}, r_{\text{agent}})$
\STATE \textit{// $\mathcal{L}$ describes expected vs actual outcome}

\STATE \textbf{// Step 2: Backward pass through DAG}
\STATE $\text{downstream} \gets \{\}$ \textit{// Accumulated downstream context}
\FOR{$v$ in $\text{ReverseTopologicalOrder}(\mathcal{G})$}
    \STATE $c_v \gets \text{GetNodeContext}(v)$ \textit{// Tool output, KB content, etc.}
    \STATE $o_v \gets \text{GetNodeOutput}(v)$ \textit{// Node's contribution to trajectory}
    \STATE $d_v \gets \text{downstream}[\text{successors}(v)]$ \textit{// How downstream used this}
    \STATE $\nabla_v \gets \text{LLM}(\textsc{GradientPrompt}, \mathcal{L}, c_v, o_v, d_v)$
    \STATE $a_v^{\text{raw}} \gets \text{ParseScore}(\nabla_v)$ \textit{// Extract [0,1] score}
    \STATE $\text{downstream}[v] \gets \nabla_v$ \textit{// Propagate gradient upstream}
\ENDFOR

\STATE \textbf{// Step 3: Normalize attribution scores}
\STATE $Z \gets \sum_{v \in V} a_v^{\text{raw}}$
\FOR{$v$ in $V$}
    \STATE $a_v \gets a_v^{\text{raw}} / Z$ \textit{// Ensure scores sum to 1.0}
\ENDFOR

\STATE \textbf{return} $\text{SortByScore}(\{(v, a_v, \nabla_v) : v \in V\})$
\end{algorithmic}
\end{algorithm}

\subsection{Illustrative Walkthrough}

Consider a simple three-node trajectory: (1)~User asks about refund policy, (2)~Agent retrieves KB entry, (3)~Agent responds with outdated 14-day policy when the correct policy is 30 days. The backward pass proceeds as follows:

\begin{itemize}
    \item \textit{Response node}: The loss is ``agent said 14 days, user expected 30 days.'' Attribution is low ($a = 0.2$) because the agent correctly used the information it was given.
    \item \textit{KB retrieval node}: The retrieved entry contained ``14-day refund policy.'' Attribution is high ($a = 0.9$) because this is where incorrect information entered the trajectory.
    \item \textit{User query node}: No attribution ($a = 0$)---the user's question was clear.
\end{itemize}

\noindent The KB entry receives the highest attribution, and its textual gradient specifies the needed update.

\subsection{Textual Gradients}

The \textit{textual gradient} $\nabla_v$ for node $v$ adapts the concept from TextGrad~\cite{textgrad2024} to conversation trajectories. Just as numerical gradients in neural network training indicate how weights should change to minimize a loss function, textual gradients provide natural language feedback describing how each component should change to improve outcomes. This analogy enables ``backpropagation through text''---propagating semantic criticism backward through the computational graph without requiring mathematical differentiability.

Concretely, the textual gradient $\nabla_v$ for node $v$ is a structured natural language description containing:
\begin{enumerate}
    \item \textbf{Attribution score} ($a_v \in [0, 1]$): A quantitative estimate of how much this component contributed to the failure, enabling prioritization across multiple potential causes.
    \item \textbf{Specific change recommendation}: What content modification would address the issue (e.g., ``add schema naming conventions to tool prompt'').
    \item \textbf{Causal justification}: Why this change would prevent the failure, linking the component's content to the observed error.
\end{enumerate}

Unlike numerical gradients that require differentiable operations, textual gradients leverage the LLM's reasoning capabilities to generate interpretable, actionable feedback for any component in the trajectory---including retrieved documents, tool descriptions, and procedural instructions that have no mathematical gradient.

\noindent\textbf{Example.} For a SQL tool that generated an invalid query, the textual gradient might be: ``\textit{Attribution: 0.85. The SQL tool description should include the constraint that all table names in this database use the `dbo\_' prefix. The agent generated `SELECT * FROM users' but the correct table name is `dbo\_users'. Adding schema naming conventions to the tool prompt would prevent this class of errors.}''

\section{DSAT Signal Taxonomy}
\label{app:taxonomy}

\begin{figure}[h]
    \centering
    \includegraphics[width=0.95\columnwidth]{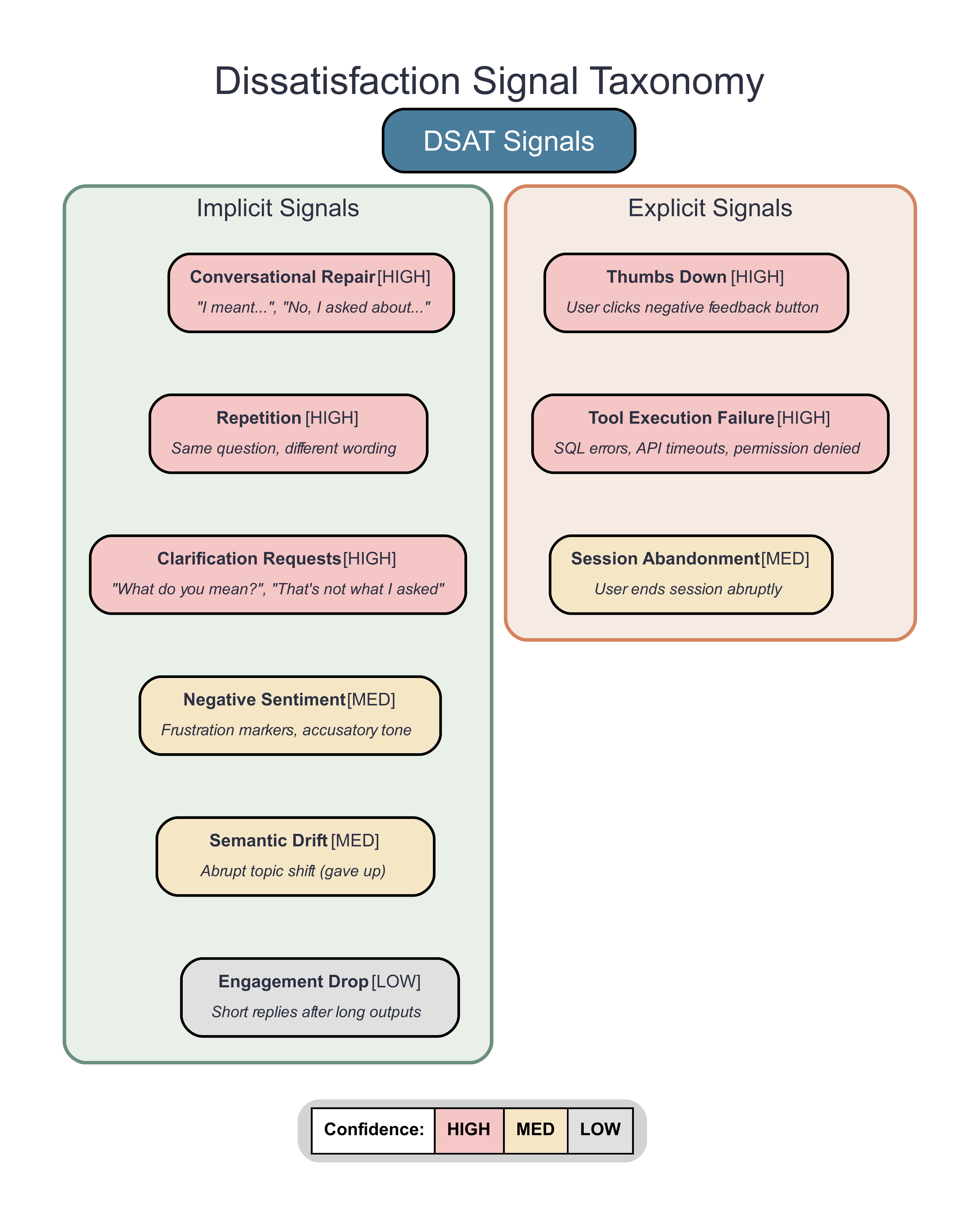}
    \caption{Taxonomy of dissatisfaction signals. Explicit signals (left) come from direct user feedback or system events. Implicit signals (right) are inferred from conversational patterns.}
    \label{fig:signals}
\end{figure}

The \detector{} identifies user dissatisfaction through two categories of signals, derived from qualitative analysis of agent conversation traces:

\textbf{Explicit Signals} (high confidence):
\begin{itemize}[nosep,leftmargin=*]
    \item \textit{Thumbs down feedback}: Direct negative feedback from UI
    \item \textit{Tool execution failure}: SQL errors, API errors, timeouts
    \item \textit{Session abandonment}: User ends session abruptly
\end{itemize}

\textbf{Implicit Signals} (require linguistic analysis):
\begin{itemize}[nosep,leftmargin=*]
    \item \textit{Conversational repair}: User rephrases or corrects (``I meant...'')
    \item \textit{Repetition}: User repeats question with different wording
    \item \textit{Negative sentiment}: Frustration markers, negative language
    \item \textit{Semantic drift}: Abrupt topic shift suggesting user gave up
\end{itemize}

Signals are weighted by confidence (explicit: $w=1.0$, high implicit: $w=0.8$, medium: $w=0.5$) and aggregated via $c = \max_{s} w_s \cdot \text{conf}_s$. Sessions with $c \geq 0.7$ or explicit signals are escalated to the \reflector{}.

\section{Prompt Templates}
\label{app:prompts}

This appendix presents the core prompts used by each \systemname{} agent.

\subsection{\detector{} System Prompt}

The \detector{} prompt embeds the signal taxonomy with examples and clear output requirements. Note that confidence aggregation is performed in post-processing code, not by the LLM:

\begin{lstlisting}[basicstyle=\ttfamily\scriptsize]
You are DetectoR, an expert at analyzing human-AI
conversations to identify user dissatisfaction signals.

## DSAT Signal Taxonomy

### 1. CORRECTION
User directly states the AI is wrong or provides the correct answer.
Examples: "That's not right, the actual value is..."
          "No, I meant X not Y" | "Wrong. It should be..."

### 2. CONVERSATIONAL_REPAIR
User redirects or repairs conversation because AI misunderstood.
Key: User is FIXING the conversation flow, not just asking for explanation.
Examples: "Let me rephrase..." | "What I actually meant was..."
Distinction: REPAIR=redirecting, CLARIFICATION=confused

### 3. TOOL_EXECUTION_ERROR (System Signal)
The AI's tool call FAILED at system level - actual execution error.
ONLY for actual system failures, NOT wrong/stale data (use CORRECTION).
Examples: SQL syntax errors, File not found, API failures
NOT tool_execution_error: Tool succeeded but returned wrong data

### 4. CLARIFICATION_REQUEST
User is CONFUSED and wants explanation, not correcting.
Examples: "What do you mean by...?" | "Can you explain that?"

### 5. NEGATIVE_SENTIMENT
User expresses frustration or disappointment.
Examples: "This isn't helpful" | "That's frustrating"

### 6. REPETITION
User repeats their original request, indicating non-fulfillment.
Examples: Restating same question | "As I said earlier..."

### 7. SEMANTIC_DRIFT
Conversation drifts from user's original intent.

### 8. SESSION_ABANDONMENT
User gives up prematurely. Examples: "Never mind" | "Forget it"

## Output Requirements
Provide structured JSON with:
- is_dsat: boolean (overall determination)
- confidence: float 0.0-1.0 (your certainty)
- signals: list of detected signals, each with:
  - signal_type: one of the 8 categories above
  - evidence: exact quote from conversation
  - confidence: float 0.0-1.0 for this signal
  - turn_id: which turn contains this signal
- trigger_turn_id: first turn showing dissatisfaction
- summary: brief explanation of findings
\end{lstlisting}

\noindent\textbf{Post-processing.} After LLM response, confidence is recomputed using weighted aggregation: $c = \max(w_s \cdot \text{conf}_s)$ where $w_s$ are predefined signal weights (system signals: 1.0, corrections: 1.0, repair/clarification/repetition: 0.8, sentiment/drift/abandonment: 0.5). Sessions with $c \geq 0.7$ or any system signal are escalated to the \reflector{}.

\subsection{\reflector{} Attribution Prompts}

The \reflector{} implements holistic attribution (Algorithm~\ref{alg:holistic}) using two coordinated prompts. We also document the iterative baseline prompts (Algorithm~\ref{alg:backward}) used in ablation studies.

\subsubsection{Stage 1: Loss Generation (The Delta)}

The first prompt extracts the \textbf{delta}---the discrepancy between what the user expected and what the agent produced. This delta serves as the loss signal for attribution:

\begin{lstlisting}[basicstyle=\ttfamily\scriptsize]
You are analyzing an AI agent conversation where user was dissatisfied.

## Conversation: {conversation}
## User Correction: {user_correction}
## DSAT Signals Detected: {dsat_signals}

## Task: Generate the DELTA (discrepancy to trace backward)

The DELTA is the gap between user expectation and agent output.
This DELTA is THE signal you will trace through the trajectory
to find the root cause node.

1. What the user EXPECTED (from their correction)
2. What the agent ACTUALLY produced
3. The DELTA: "Expected X but got Y" (one sentence)

Output: {"loss_description": "DELTA: Expected [X] but got [Y]",
         "expected_outcome": "...", "actual_outcome": "..."}
\end{lstlisting}

\subsubsection{Stage 2: Holistic Attribution (Primary Method)}

The holistic approach presents the agent's execution \emph{trace} to the LLM---chain-of-thought, tool calls, and tool outputs---in reverse temporal order. The LLM does not separately inspect raw context files; it relies on the chain-of-thought, where the agent itself names the sources that shaped each decision. Verification of those candidate files is deferred to the \recommender{} stage. This prompts a single-pass backward walk in the spirit of gradient descent's backward pass through a computational graph:

\begin{lstlisting}[basicstyle=\ttfamily\scriptsize]
You are performing holistic root cause attribution.

## SCOPE (Important)
You see the agent's execution TRACE only: chain-of-thought,
tool calls, and tool outputs. You do NOT have the raw KB, the
full set of Skill files, or every tool definition.

The chain-of-thought is your most diagnostic signal -- the
agent's own reasoning names the sources behind each decision
(e.g., "According to the KB..." or "Following the SOP for...").
Use those references to identify candidate root-cause sources;
verification of file contents is the Recommender's job.

## THE DELTA (Your Search Target)
Loss: {loss_description}
Expected: {expected_outcome} | Actual: {actual_outcome}

## TRAJECTORY (Agent-Accessed Sources, REVERSE Temporal Order)
Listed from the agent's final response backward to the
original user query. Each entry contains the node's content
(input it received) and output (what it produced/decided).

{trajectory_nodes_in_reverse_order_with_content_and_output}

## TASK: Identify Root Cause via Reverse-Order Backward Pass
Treat the trajectory as a computational graph and the DELTA
as the loss signal. Walk the trajectory in REVERSE TEMPORAL
ORDER -- gradient-descent style -- to localize the fault:

  1. START at the agent's FINAL RESPONSE (where loss is observed).
  2. Step BACKWARD one node at a time:
       response -> tool outputs -> KB retrievals -> skill lookups
                -> tool prompts -> system prompt -> user query.
     For each node ask: "Does the DELTA already appear in this
     node's OUTPUT? Did this node's INPUT already contain it?"
  3. The EARLIEST node whose OUTPUT contains the DELTA but whose
     INPUT does NOT is the node that INTRODUCED the error
     (the root cause). All later nodes merely PROPAGATED it.
  4. Use the simultaneous view to disambiguate ties: among
     candidate root-cause nodes, pick the one whose content is
     most directly inconsistent with the DELTA.

This mirrors textual gradient backpropagation: the DELTA is the
loss; backward traversal is the gradient signal; the root-cause
node is where the gradient is largest -- the "weight" most
responsible for the loss.

NOTE: If the true root cause is a context source the agent
NEVER ACCESSED (e.g., a missing KB entry, an SOP the agent
should have read but did not), the trajectory cannot directly
reveal it. In that case, attribute to the EARLIEST accessed
node whose decision shows the agent should have looked further
(retrieval failure, missing-skill lookup, etc.), and let the
RecommendeR's exploration stage verify and refine.

## Four-Dimensional Diagnosis (for root cause node)
- existence: Does required information exist?
- accessibility: Was it successfully retrieved?
- correctness: Is the information accurate and up-to-date?
- consistency: Is it consistent with other sources?

## Output Schema
{
  "root_cause_node": "<node_id>",
  "attribution_scores": {"node_id": <float 0.0-1.0>, ...},
  "fault_category": "<category>",
  "component_path": "<path to fix>",
  "gradient_description": "<what should change>",
  "diagnosis": {existence, accessibility, correctness, consistency}
}
\end{lstlisting}

\subsubsection{Iterative Per-Node Attribution (Ablation Baseline)}

For ablation comparison, we also implement iterative per-node attribution. For each node, the prompt asks: \textit{Does this node's output contain the delta?} If the delta appears in the output but not the input, this node \textit{introduced} the error. The \textbf{Error Transformation Classification} formalizes this logic:

\begin{lstlisting}[basicstyle=\ttfamily\scriptsize]
You are performing per-node attribution analysis (ablation baseline).

## THE DELTA (Your Search Target)
Loss: {loss_description}
Expected: {expected_outcome} | Actual: {actual_outcome}

QUESTION: Does THIS NODE's output contain the DELTA?
- If YES and NOT in input -> This node INTRODUCED the error
- If YES and also in input -> This node PROPAGATED the error
- If NO -> This node is not involved

## Full Trajectory Context (all nodes for reference)
{full_trajectory_context}

## FOCUS: Analyze Node {node_id}
- Type: {node_type}
- Full Input: {node_input}
- Full Output: {node_output}
- Reasoning: {node_reasoning}
- Tool Info: {tool_info}

## Downstream Evidence (Observable Facts from Later Nodes)
{downstream_evidence}

## CRITICAL: Error Transformation Classification

Classify how this node relates to the error (input vs output):

INTRODUCED: Input is CORRECT, output is INCORRECT
  - This node is where the error ORIGINATED
  - High attribution (0.7-0.9)

PROPAGATED: Input already has error, output passes it unchanged
  - This node just passed along an existing error
  - Low attribution (0.1-0.2)

AMPLIFIED: Input has minor issue, output makes it worse
  - This node worsened an existing problem
  - Medium-high attribution (0.5-0.7)

TRANSFORMED: Input has error type A, output has error type B
  - This node changed the nature of the error
  - Medium attribution (0.3-0.5)

NONE: Neither input nor output contains relevant error
  - This node is unrelated to the failure
  - No attribution (0.0-0.1)

## Four-Dimensional Diagnosis
- existence: Does required information exist in knowledge source?
- accessibility: Was it successfully retrieved/accessed?
- correctness: Is the information accurate and up-to-date?
- consistency: Is it consistent with other sources?

## Output Schema
{
  "node_id": {node_id},
  "error_analysis": {
    "error_present_in_input": true/false,
    "error_present_in_output": true/false,
    "transformation_type": "introduced|propagated|...|none",
    "error_signature": "<specific error pattern found>",
    "evidence": "<data supporting classification>"
  },
  "attribution_score": <float 0.0-1.0>,
  "component_type": "<skill|kb_entry|tool_prompt|system_prompt|none>",
  "component_path": "<path if applicable>",
  "gradient_description": "<what should change to fix this?>",
  "diagnosis": {existence, accessibility, correctness, consistency},
  "is_root_cause_candidate": true/false,
  "cascade_from": <upstream node id if error propagated, else null>
}
\end{lstlisting}

\subsection{\recommender{} Agentic Exploration Prompt}

The \recommender{} is an agentic LLM with tool access that explores context sources to verify the \reflector{}'s hypothesis. Unlike static prompt chains, it autonomously decides which tools to call based on what it discovers. The system prompt establishes the exploration protocol and available tools:

\begin{lstlisting}[basicstyle=\ttfamily\scriptsize]
You are the Recommender Agent in the TRACE system. Your task is to
VERIFY the Root Cause Agent's hypothesis through ACTIVE EXPLORATION
of the agent's context sources, then generate a CRUD recommendation.

## Your Mission
The Root Cause Agent identified a potential failure source. You must:
1. EXPLORE: Use tools to investigate the implicated component
2. VERIFY: Cross-reference against authoritative sources
3. DIAGNOSE: Apply four-dimensional analysis
4. RECOMMEND: Generate actionable CRUD recommendation

## Available Tools (USE THEM!)
- search_kb(query): Search knowledge base for related content
- read_kb_entry(id): Read full KB entry with metadata
- list_skills(path): Browse Skills file hierarchy
- read_skill(path): Read Skill/SOP document content
- read_system_prompt(): Get current system prompt
- read_tool_prompt(name): Get tool description

## Exploration Protocol
1. READ the implicated component (get current state, metadata)
2. SEARCH for authoritative sources (policies, specifications)
3. CROSS-REFERENCE implicated content vs authoritative sources
4. EXPLORE related components that may need coordinated updates

## Four-Dimensional Diagnosis (evaluate ALL dimensions)
D1-EXISTENCE: Does content exist? (Gap = CREATE)
D2-ACCESSIBILITY: Was it retrieved? (Routing fail = UPDATE routing)
D3-CORRECTNESS: Is it accurate? (Stale/wrong = UPDATE content)
D4-CONSISTENCY: Aligned with other sources? (Conflict = UPDATE)

## Authority Resolution (when sources conflict)
1. Explicit supersession ("replaces policy X")
2. Official status (CFO-approved > draft)
3. Hierarchy (system-level > component-level)
4. Recency (newer > older)
5. Domain ownership

## Classification
SYSTEMIC: Verified problem affecting many users -> HIGH priority
INCIDENTAL: No issue found, edge case -> NO_ACTION
AMBIGUOUS: Cannot verify -> LOW priority, flag for human

## Input
Root Cause Analysis: {root_cause_analysis}
Implicated Component: {component_path}
Fault Category: {fault_category}
User Correction: {user_correction}

## Output Schema
After exploration, output:
{
  "exploration_log": [{"tool": "...", "result": "..."}],
  "diagnosis": {
    "existence": {"passed": bool, "evidence": "..."},
    "accessibility": {"passed": bool, "evidence": "..."},
    "correctness": {"passed": bool, "evidence": "..."},
    "consistency": {"passed": bool, "evidence": "..."}
  },
  "operation": "CREATE|UPDATE|DELETE|NO_ACTION",
  "target_path": "path/to/fix",
  "recommended_change": "specific change text",
  "authoritative_source": "source used for validation",
  "classification": "SYSTEMIC|INCIDENTAL|AMBIGUOUS",
  "confidence": 0.0-1.0
}
\end{lstlisting}

\noindent The agent typically makes 3--5 tool calls per trace, exploring the implicated component, searching for authoritative sources, and checking for related content that may need coordinated updates.

\section{Evaluation Metric Definitions}
\label{app:metrics}

This appendix formally defines the evaluation metrics used throughout the paper.

\subsection{\detector{} Metrics}

We evaluate binary DSAT detection using standard classification metrics: precision, recall, and F1 score.

\subsection{\reflector{} Agent Metrics}

\textbf{Node Accuracy (Acc@1):} Whether the predicted root cause node matches ground truth:
\begin{equation}
\text{Acc@1} = \frac{1}{N} \sum_{i=1}^{N} \mathbf{1}[\hat{n}_i = n_i^*]
\end{equation}
where $\hat{n}_i$ is the predicted root cause node and $n_i^*$ is the ground truth.

\textbf{Accuracy@3 (Acc@3):} Whether the ground truth node is among the top-3 attributed nodes:
\begin{equation}
\text{Acc@3} = \frac{1}{N} \sum_{i=1}^{N} \mathbf{1}[n_i^* \in \text{top-3}(\hat{A}_i)]
\end{equation}
where $\hat{A}_i$ is the predicted attribution score vector.

\textbf{Category Accuracy:} Whether the predicted fault category matches ground truth:
\begin{equation}
\text{Cat. Acc} = \frac{1}{N} \sum_{i=1}^{N} \mathbf{1}[\hat{c}_i = c_i^*]
\end{equation}

\subsection{\recommender{} Metrics}

\textbf{CRUD Accuracy:} Whether the predicted CRUD operation matches ground truth:
\begin{equation}
\text{CRUD Acc} = \frac{1}{N} \sum_{i=1}^{N} \mathbf{1}[\hat{o}_i = o_i^*]
\end{equation}
where $\hat{o}_i \in \{\text{CREATE}, \text{UPDATE}, \text{DELETE}, \text{NO\_ACTION}\}$.

\textbf{Path Match (Exact):} Whether the target path exactly matches ground truth:
\begin{equation}
\text{Path (Exact)} = \frac{1}{N} \sum_{i=1}^{N} \mathbf{1}[\hat{p}_i = p_i^*]
\end{equation}

\textbf{Path Match (Partial):} Whether the target path contains the correct component:
\begin{equation}
\text{Path (Partial)} = \frac{1}{N} \sum_{i=1}^{N} \mathbf{1}[\text{component}(\hat{p}_i) = \text{component}(p_i^*)]
\end{equation}

A partial match occurs when the predicted path references the correct file but may differ in exact formatting (e.g., ``skills/vendor-payment/SKILL.md'' vs.\ ``vendor-payment'').

\subsection{Ablation Study Metrics}

\textbf{Recovery Rate:} The proportion of initially incorrect predictions that are corrected through exploration:
\begin{equation}
\text{Recovery Rate} = \frac{|\{i : \hat{n}_i^{(1)} \neq n_i^* \land \hat{n}_i^{(2)} = n_i^*\}|}{|\{i : \hat{n}_i^{(1)} \neq n_i^*\}|}
\end{equation}
where $\hat{n}_i^{(1)}$ is the initial prediction and $\hat{n}_i^{(2)}$ is the post-exploration prediction.

\textbf{LLM Call Efficiency:} The average number of LLM calls per trace:
\begin{equation}
\text{Calls/Trace} = \frac{1}{N} \sum_{i=1}^{N} \text{calls}_i
\end{equation}

This metric captures the compute cost trade-off between iterative (multiple calls) and holistic (single call) approaches.

\subsection{End-to-End Pipeline Metrics}

\textbf{Cumulative Accuracy:} The proportion of traces where all three agents produce correct outputs:
\begin{equation}
\text{E2E Acc} = \frac{1}{N} \sum_{i=1}^{N} \mathbf{1}[\text{DSAT}_i \land \text{Node}_i \land \text{CRUD}_i]
\end{equation}

This metric captures the compounding effect of errors across the pipeline.


\begin{thebibliography}{99}

\bibitem{textgrad2024}
Mert Yuksekgonul, Federico Bianchi, Joseph Boen, Sheng Liu, Zhi Huang, Carlos Guestrin, and James Zou.
\newblock TextGrad: Automatic ``Differentiation'' via Text.
\newblock \emph{arXiv preprint arXiv:2406.07496}, 2024.

\bibitem{zhou2024semantic}
Wenyi Wang, Hisham A. Alyahya, Dylan R. Ashley, Oleg Serikov, Dmitrii Khizbullin, Francesco Faccio, and J\"urgen Schmidhuber.
\newblock How to Correctly do Semantic Backpropagation on Language-based Agentic Systems.
\newblock \emph{arXiv preprint arXiv:2412.03624}, 2024.

\bibitem{drift2025}
Yifan Wang, Bolian Li, Junlin Wu, Zhaoxuan Tan, Zheli Liu, Ruqi Zhang, Ananth Grama, and Qingkai Zeng.
\newblock DRIFT: Learning from Abundant User Dissatisfaction in Real-World Preference Learning.
\newblock \emph{arXiv preprint arXiv:2510.02341}, 2025.

\bibitem{ace2024}
Qizheng Zhang, Changran Hu, Shubhangi Upasani, Boyuan Ma, Fenglu Hong, Vamsidhar Kamanuru, Jay Rainton, Chen Wu, Mengmeng Ji, Hanchen Li, Urmish Thakker, James Zou, and Kunle Olukotun.
\newblock Agentic Context Engineering: Evolving Contexts for Self-Improving Language Models.
\newblock \emph{arXiv preprint arXiv:2510.04618}, 2025.

\bibitem{protegi2023}
Reid Pryzant, Dan Iter, Jerry Li, Yin Tat Lee, Chenguang Zhu, and Michael Zeng.
\newblock Automatic Prompt Optimization with ``Gradient Descent'' and Beam Search.
\newblock In \emph{Proceedings of EMNLP}, 2023.

\bibitem{crispo2025}
Han He, Qianchu Liu, Lei Xu, Chaitanya Shivade, Yi Zhang, Sundararajan Srinivasan, and Katrin Kirchhoff.
\newblock CriSPO: Multi-Aspect Critique-Suggestion-guided Automatic Prompt Optimization for Text Generation.
\newblock In \emph{Proceedings of AAAI}, 2025.

\bibitem{pace2023}
Yihong Dong, Kangcheng Luo, Xue Jiang, Zhi Jin, and Ge Li.
\newblock PACE: Improving Prompt with Actor-Critic Editing for Large Language Model.
\newblock \emph{arXiv preprint arXiv:2308.10088}, 2023.

\bibitem{icai2025}
Arduin Findeis, Timo Kaufmann, Eyke H\"ullermeier, Samuel Albanie, and Robert D. Mullins.
\newblock Inverse Constitutional AI: Compressing Preferences into Principles.
\newblock In \emph{Proceedings of ICLR}, 2025.

\bibitem{mem02025}
Prateek Chhikara, Dev Khant, Saket Aryan, Taranjeet Singh, and Deshraj Yadav.
\newblock Mem0: Building Production-Ready AI Agents with Scalable Long-Term Memory.
\newblock \emph{arXiv preprint arXiv:2504.19413}, 2025.

\bibitem{amber2025}
Qitao Qin, Yucong Luo, Yihang Lu, Zhibo Chu, Xiaoman Liu, and Xianwei Meng.
\newblock Towards Adaptive Memory-Based Optimization for Enhanced Retrieval-Augmented Generation.
\newblock In \emph{Findings of ACL}, 2025.

\bibitem{llmagentsurvey2025}
Zhipeng Liu, Xuefeng Bai, Kehai Chen, Xinyang Chen, Xiucheng Li, Yang Xiang, Jin Liu, Hong-Dong Li, Yaowei Wang, Liqiang Nie, and Min Zhang.
\newblock A Survey on the Feedback Mechanism of LLM-based AI Agents.
\newblock In \emph{Proceedings of IJCAI}, 2025.

\bibitem{rlhf2022}
Long Ouyang, Jeff Wu, Xu Jiang, Diogo Almeida, Carroll L. Wainwright, Pamela Mishkin, Chong Zhang, Sandhini Agarwal, Katarina Slama, Alex Ray, et al.
\newblock Training language models to follow instructions with human feedback.
\newblock In \emph{Advances in Neural Information Processing Systems}, 2022.

\bibitem{multiagent2024}
Wanjia Zhao, Mert Yuksekgonul, Shirley Wu, and James Zou.
\newblock SiriuS: Self-improving Multi-agent Systems via Bootstrapped Reasoning.
\newblock \emph{arXiv preprint arXiv:2502.04780}, 2025.

\bibitem{liu2023agentbench}
Xiao Liu, Hao Yu, Hanchen Zhang, Yifan Xu, Xuanyu Lei, Hanyu Lai, Yu Gu, Hangliang Ding, Kaiwen Men, Kejuan Yang, Shudan Zhang, Xiang Deng, Aohan Zeng, Zhengxiao Du, Chenhui Zhang, Sheng Shen, Tianjun Zhang, Yu Su, Huan Sun, Minlie Huang, Yuxiao Dong, and Jie Tang.
\newblock AgentBench: Evaluating LLMs as Agents.
\newblock In \emph{Proceedings of ICLR}, 2024.

\bibitem{mialon2023gaia}
Gr\'egoire Mialon, Cl\'ementine Fourrier, Craig Swift, Thomas Wolf, Yann LeCun, and Thomas Scialom.
\newblock GAIA: A Benchmark for General AI Assistants.
\newblock \emph{arXiv preprint arXiv:2311.12983}, 2023.

\bibitem{chen2024benchmarking}
Jiawei Chen, Hongyu Lin, Xianpei Han, and Le Sun.
\newblock Benchmarking Large Language Models in Retrieval-Augmented Generation.
\newblock In \emph{Proceedings of AAAI}, 2024.

\end{thebibliography}
\end{document}